\documentclass[final,5p,times,twocolumn]{elsarticle}

\usepackage{graphicx}

\usepackage{amssymb,amsmath}
\usepackage{color,psfrag}
\usepackage{algorithm,algorithmic}
\usepackage{lscape}
\usepackage{tikz}
\usepackage{colortbl}

\definecolor{ncol}{rgb}{0.00,0.00,0.00}
\definecolor{bcol}{rgb}{0.90,0.50,0.00}
\definecolor{pcol}{rgb}{0.30,0.15,0.75}

\definecolor{cg}{rgb}{0.00,0.00,0.00}
\definecolor{cd}{rgb}{1.00,0.25,0.15}
\definecolor{cw}{rgb}{0.90,0.75,0.10}

\tikzstyle{ng} = [color=ncol]
\tikzstyle{nd} = [color=ncol]
\tikzstyle{nw} = [color=ncol]
\tikzstyle{pg} = [color=pcol]
\tikzstyle{pd} = [color=pcol]
\tikzstyle{pw} = [color=pcol]
\tikzstyle{bg} = [color=bcol]
\tikzstyle{bd} = [color=bcol]
\tikzstyle{bw} = [color=bcol]

\tikzstyle{gst} = [thick]
\tikzstyle{dst} = [thick,dotted]
\tikzstyle{wst} = [thick,dashed]

\tikzstyle{nge} = [gst,color=ncol]
\tikzstyle{nde} = [dst,color=ncol]
\tikzstyle{nwe} = [wst,color=ncol]
\tikzstyle{pge} = [gst,color=pcol]
\tikzstyle{pde} = [dst,color=pcol]
\tikzstyle{pwe} = [wst,color=pcol]
\tikzstyle{bge} = [gst,color=bcol]
\tikzstyle{bde} = [dst,color=bcol]
\tikzstyle{bwe} = [wst,color=bcol]

\newcommand{\tick}[1]{\scriptsize{#1}}
\newcommand{\sub}[1]{$_{\text{#1}}$}

\newcommand{\best}[1]{\colorbox[gray]{0.10}{\color{white} #1}}
\newcommand{\ten}[1]{\colorbox[gray]{0.60}{#1}}
\newcommand{\forty}[1]{\colorbox[gray]{0.85}{#1}}

\newcommand{\cfill}{\cellcolor[gray]{0.9}}

\newcommand{\plotlegend}{%
  \def\legendborderleft{0.5}
  \def\legendborderright{1.5}
  \def\legendbordertop{0.2}
  \def\legendborderbottom{0.2}
  \def\hrow{0.3}
  \begin{tikzpicture}[xscale=1,yscale=-1]
    \node (leg1) at (3,45) {};
    \path (leg1) +(2.3,0) node (leg2) {};
    \path (leg2) +(2.2,0) node (leg3) {};

    \path (leg1.north west)+(-\legendborderleft,-\legendbordertop)
      node (a) {};
    \path (leg3.south east)+(\legendborderright,\legendborderbottom)
      node (b) {};

    \path[draw=black,fill=gray!10] (a) rectangle (b);    

    \node[anchor=west,font=\scriptsize] at (leg1) {NOISE};
    \node[anchor=west,font=\scriptsize] at (leg2) {POUT};
    \node[anchor=west,font=\scriptsize] at (leg3) {BALL};
    
    \node[rectangle,fill=ncol,inner sep=2pt] at (leg1.west) {};
    \node[rectangle,fill=pcol,inner sep=2pt] at (leg2.west) {};
    \node[rectangle,fill=bcol,inner sep=2pt] at (leg3.west) {};
    
  \end{tikzpicture}
}

\newcommand{\plotlegendextra}{%
  \def\legendborderleft{0.5}
  \def\legendborderright{1.5}
  \def\legendbordertop{0.2}
  \def\legendborderbottom{0.2}
  \def\hrow{0.3}
  \begin{tikzpicture}[xscale=1,yscale=-1]
    \node (leg1) at (3,45) {};
    \path (leg1) +(2.3,0) node (leg2) {};
    \path (leg2) +(2.2,0) node (leg3) {};

    \path (leg1) +(0,\hrow) node (leg1b) {};
    \path (leg2) +(0,\hrow) node (leg2b) {};
    \path (leg3) +(0,\hrow) node (leg3b) {};
  
    \path (leg1.north west)+(-\legendborderleft,-\legendbordertop)
      node (a) {};
    \path (leg3b.south east)+(\legendborderright,\legendborderbottom)
      node (b) {};

    \path[draw=black,fill=gray!10] (a) rectangle (b);    

    \node[anchor=west,font=\scriptsize] at (leg1) {NOISE};
    \node[anchor=west,font=\scriptsize] at (leg2) {POUT};
    \node[anchor=west,font=\scriptsize] at (leg3) {BALL};
    
    \node[rectangle,fill=ncol,inner sep=2pt] at (leg1.west) {};
    \node[rectangle,fill=pcol,inner sep=2pt] at (leg2.west) {};
    \node[rectangle,fill=bcol,inner sep=2pt] at (leg3.west) {};
    
    \node[anchor=west,font=\scriptsize] at (leg1b) {GRAYMAT};
    \node[anchor=west,font=\scriptsize] at (leg2b) {DOCS};
    \node[anchor=west,font=\scriptsize] at (leg3b) {WDOCS};
    
    \draw[gst] ($(leg1b.west)+(-0.2,0)$) -- +(0.4,0);
    \draw[dst] ($(leg2b.west)+(-0.2,0)$) -- +(0.4,0);
    \draw[wst] ($(leg3b.west)+(-0.2,0)$) -- +(0.4,0);
    
  \end{tikzpicture}
}

\newcommand{\comment}[1]{}

\begin{document}

\begin{frontmatter}



\title{Algorithms for Grey-Weighted Distance Computations}


\author{M. Gedda\corref{cor1}}
\cortext[cor1]{Tel.: +46~18~471~7849; fax: +46~18~553447\\E-mail: {\tt magnus.gedda@cb.uu.se}}
\address{Centre for Image Analysis, Uppsala University, Box 337, SE-751 05, Uppsala, Sweden}

\begin{abstract}
With the increasing size of datasets and demand for real time response for interactive applications, improving runtime for algorithms with excessive computational requirements has become increasingly important. Many different algorithms combining efficient priority queues with various helper structures have been proposed for computing grey-weighted distance transforms. Here we compare the performance of popular competitive algorithms in different scenarios to form practical guidelines easy to adopt. The label-setting category of algorithms is shown to be the best choice for all scenarios. The hierarchical heap with a pointer array to keep track of nodes on the heap is shown to be the best choice as priority queue. However, if memory is a critical issue, then the best choice is the Dial priority queue for integer valued costs and the Untidy priority queue for real valued costs.
\end{abstract}


\end{frontmatter}


\usetikzlibrary{shapes,arrows,calc}

\section{Introduction}\label{sec:introduction}

Image analysis measurements are generally performed on binary representations of the objects. However, when images are acquired, grey levels have specific meanings. Binarisation of such images results in a loss of information and neither the internal intensities nor the borders of the resulting regions represent the imaged objects very well. This can be due to limited resolution, high noise levels, or that the border is a compound of objects. Because of this, measurements are increasingly done directly on grey-level images~\cite{mg:SladojePAMI2009}. Fuzzy theory~\cite{mg:fuzzy:ZadehIC1965}, where an image element has a membership value describing its belongingness to a certain (fuzzy) object, has emerged as a framework for addressing these problems~\cite{mg:CarvalhoPAA1999,mg:UdupaGMIP1996}.


Distance calculations are widely used to extract shape and size information~\cite{mg:BorgeforsAVFP1994,mg:JonesTVCG2006}. This is an area where measurements in grey-level images have become increasingly popular~\cite{mg:BlochPR2000,mg:GRAYMAT:LeviIC1970,mg:PreteuxSPIE1991,mg:GWD:RutovitzPPR1968,mg:FDT:SahaCVIU2002,mg:SoillePRL1994,mg:ToivanenPRL1996}. The applications for content-based distance measures are many, e.g., grey-level morphology and minimal path detection~\cite{mg:SoillePRL1994}, segmentation~\cite{mg:MeyerSP1994}, clustering~\cite{mg:GeddaIWCIA2006}, and solving the Eikonal equation~\cite{mg:KimmelJMIV1996,mg:Sethian1996,mg:VerbeekPRL1990}. With the expanding size of datasets and demand for real time response for both automatic and interactive applications, improving memory efficiency and runtime for algorithms with excessive computational requirements has become a focus of greater importance~\cite{mg:CotoCG2007,mg:MalmbergDGCI2006,mg:NyulGM2003}. Due to hardware limitations, the first efficient methods for computing distance transforms were based on the classic raster scan approach~\cite{mg:SeqOp:RosenfeldJACM1966}. This approach works well for distance calculations on binary images, where a complete distance transform only needs two passes through the image. But for grey-level images, where the domain is generally not convex, the number of passes through the image becomes dependent on content. To improve runtime, propagation using graph-search techniques have become popular~\cite{mg:FalcaoTMI2000,mg:IkonenDGCI2005,mg:VerbeekPRL1990}. Most of the methods are versions of the well-known, theoretically optimal Dijkstra's algorithm~\cite{mg:DijkstraNM1959}.
The wealth of data structures available for these algorithms makes analysing the computational complexity of all different combinations nontrivial. Even if it was trivial, implementations of lowest complexity might still not be the fastest due to practical implications. For example, the work by Luengo~\cite{mg:LuengoIJCV2009} shows the impact of current computer hardware on different priority queues. Also, special situations that often arise in image analysis problems, such as spatial homogeneity in images or overhead of complex structures when working on small problem domains, can also be a factor to why implementations with higher complexity might perform better than ones with low complexity.

Kimmel et al.~\cite{mg:KimmelJMIV1996} calculated the grey-weighted distance by solving the Eikonal equation using the Fast Marching method (FMM)~\cite{mg:Sethian1996}, which is an efficient numerical scheme for solving the continuous boundary value problems. Here we focus on {\em discrete} distance definitions, covered in Section~\ref{sec:principles}, and compare algorithms aimed to find the shortest path in a network with prescribed weights for each link between nodes. Numerical methods for approximating the solutions of a continuous problems are out of scope of this paper. 

We put different implementations of the most popular grey-weighted distance transform algorithms, which we cover in Section~\ref{sec:algorithms}, in a comparative test, under settings representative of common situations in image analysis, in Section~\ref{sec:experiments}. The work by Nyul et al.~\cite{mg:NyulGM2003} presents a similar study on algorithms for fuzzy-connected image segmentation. However, it is important to point out that the results do not apply to grey-weighted distance transforms due to the different properties of fuzzy connectedness. Since our work relates to the same subject we have chosen to use similar terminology. We incorporate all algorithms and data structures used by Nyul et al.~\cite{mg:NyulGM2003} and also include the data structures introduced by Yatziv et al.~\cite{mg:YatzivJCP2006} and Luengo~\cite{mg:LuengoIJCV2009}. The conclusions in Section~\ref{sec:conclusion} should be seen as practical guidelines for selecting grey-weighted distance algorithms in different scenarios. To keep the adherence of the guidelines from being overly complex and off-putting, we have chosen competitive algorithms of low programming complexity and use containers from the C++ Standard Template Library (STL)~\cite{mg:Josuttis1999} where applicable. We focus on local sequential algorithms, all parallel algorithms are out of scope for this article.


\section{Discrete grey-weighted distances}
\label{sec:principles}

The geodesic distance between two points included in a set is the length of the shortest paths or geodesics~\cite{mg:SerraIAMM1988} linking these points and included in the set. The set is referred to as a {\it geodesic mask}, and when calculating grey-weighted distance the grey-scale geodesic mask is usually the same as the input image. In this paper we define a discrete grey-scale image $f : \mathbb{Z}^n \longrightarrow \mathbb{R}^*$ as an application of a subset of the $n$-dimensional discrete space $\mathbb{Z}^n$ into the set $\mathbb{R}^*$ of non-negative real numbers. The neighbourhood relations between the points in a discrete image are defined by a graph. We use an 8-connected graph for 2D square grids, and a 26-connected graph for 3D cubic grids. We define a discrete path $P$ of length $l - 1$ going from node $p$ to node $q$ as a $l$-tuple $(x_1,\ldots,x_l)$ of nodes such that $x_1 = p$, $x_l = q$, and $(x_i,x_{i+1})$ defines adjacent nodes for all $i = 1,\ldots,l-1$. The grey-weighted distance $d(p,q)$ represents the sum of all arc weights $c_i$ along $P$. This assumes that the arc weight $c_i$ represents the cost of travelling from a node $x_i$ to node $x_{i+1}$. The grey-weighted distance $d(p,q)$ then consists of finding the path with the lowest sum of arc weights $c_i$ along all possible paths linking $p$ to $q$. If the set $\mathcal{P}_{pq}$ consists of all possible paths from $p$ to $q$, we have
\[
d(p,q) = \{ \, \min_{P\in\mathcal{P}_{pq}} \, (\mathcal{C}(P)) \,\, | \,\, \mathcal{C}(P) = \sum_{i}^{l-1}c_i(x_i,x_{i+1}) \,\, \},
\]
where the arc weight $c_i$, also referred to as {\it cost} or {\it local cost}, is calculated by a cost function on the geodesic mask $f$. Although the definition is general for $n$ dimensions we refer to image elements as pixels (or nodes when utilising the graph analogy) unless we operate on 3D images, where we refer to them as voxels. 

Rutovitz first proposed a grey-weighted distance where the arc weight is equal to the grey level of the destination pixel of each step along the path~\cite{mg:GWD:RutovitzPPR1968}. Levi and Montanari extended this definition when they defined a grey-weighted medial axis transform (GRAYMAT) by weighting the grey levels with the distance between adjacent pixels along the path~\cite{mg:GRAYMAT:LeviIC1970}. In their definition, the length of a path is defined as the discretisation of the integral of the pixel values along the path, and the arc weight is defined as
\begin{equation}\label{eq:graymat}
c_i = \frac{1}{2}(f(x_i) + f(x_{i+1})) \cdot ||x_i - x_{i+1}||,
\end{equation}
where $||\cdot||$ refers to the spatial distance between two adjacent nodes in the image graph.
Saha et al. proposed a theoretical framework for the GRAYMAT definition when applied to fuzzy sets~\cite{mg:FDT:SahaCVIU2002}. Soille also defined a geodesic measure for fuzzy sets inspired by Levi and Montanari's definition~\cite{mg:SoillePRL1994}. For more distance definitions on fuzzy sets we refer to~\cite{mg:BlochPR1999}.

Toivanen proposed two definitions for arc weights where the path between two points is defined as a path lying on the hyperplane defined by the grey levels~\cite{mg:ToivanenPRL1996}. The first is the distance on curved space (DOCS),
\begin{equation}\label{eq:docs}
c_i = |f(x_i) - f(x_{i+1})| + ||x_i - x_{i+1}||,
\end{equation}
and the second is the weighted distance on curved space (WDOCS),
\begin{equation}\label{eq:wdocs}
c_i = \sqrt{|f(x_i) - f(x_{i+1})|^2 + ||x_i - x_{i+1}||^2}.
\end{equation}
While GRAYMAT propagates fast for low grey levels, DOCS and WDOCS account for the changes in height of the 'height map' and represent the minimal amount of ascents and descents to be travelled to reach a neighbouring pixel. DOCS performs the distance calculation with integer numbers while each sub-distance along the path for WDOCS is euclidean.

\begin{figure*}
  \newcommand{\annot}{%
    \clip (0,0) rectangle (127,127);
    \path[draw=black,line width=2pt] (0,100) -- (56,100);
    \path[draw=red,line width=1pt] (0,100) -- (55.5,100);
    \path[draw=black,line width=2pt] (71,100) -- (127,100);
    \path[draw=red,line width=1pt] (71.5,100) -- (127,100);

    \path[draw=black,line width=2pt] (19,24) -- (19,75) -- (108,75) -- (108,24);
    \path[draw=red,line width=1pt] (19,24.5) -- (19,75) -- (108,75) -- (108,24.5);

    \path[draw=black,line width=1.5pt] (63.2,111.2)
    +(45:4) -- +(225:4) +(-45:4) -- +(-225:4);
    \path[draw=green,line width=0.8pt] (63.2,111.2)
    +(45:3.5) -- +(225:3.5) +(-45:3.5) -- +(-225:3.5);
  }

  \newlength\imagewidth
  \newlength\imagescale
  \pgfmathsetlength{\imagewidth}{3cm} 
  \def\imagescale{0.0234375}
\begin{minipage}[b]{0.10\linewidth}
  \centerline{}
\end{minipage}
\hfill
\begin{minipage}[b]{0.18\linewidth}
  \begin{center}
  \begin{tikzpicture}[xscale=\imagescale,yscale=-\imagescale]
    \node[anchor=north west,inner sep=0pt,outer sep=0pt] at (-0.5,-0.5)
       {\includegraphics[width=\imagewidth]{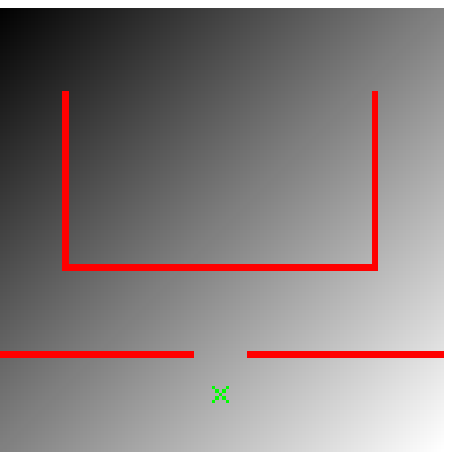}};
    \annot
  \end{tikzpicture}
  \end{center}
  \vspace{0.1cm}
\end{minipage}
\hfill
\begin{minipage}[b]{0.18\linewidth}
  \begin{center}
  \begin{tikzpicture}[xscale=\imagescale,yscale=-\imagescale]
    \node[anchor=north west,inner sep=0pt,outer sep=0pt] at (-0.5,-0.5)
       {\includegraphics[width=\imagewidth]{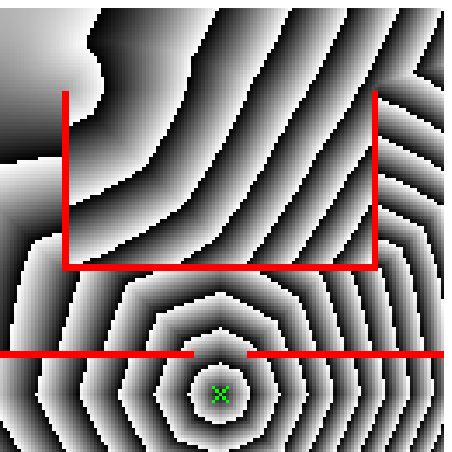}};
    \annot
  \end{tikzpicture}
  \end{center}
  \vspace{0.1cm}
\end{minipage}
\hfill
\begin{minipage}[b]{0.18\linewidth}
  \begin{center}
  \begin{tikzpicture}[xscale=\imagescale,yscale=-\imagescale]
    \node[anchor=north west,inner sep=0pt,outer sep=0pt] at (-0.5,-0.5)
       {\includegraphics[width=\imagewidth]{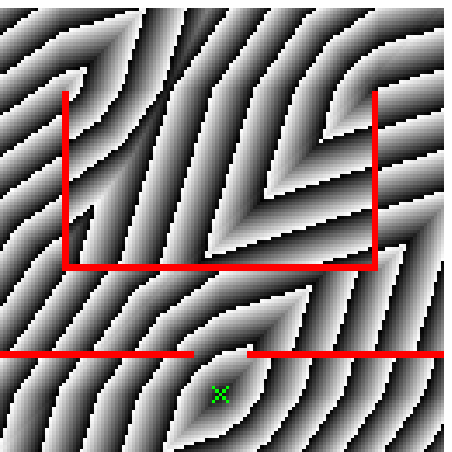}};
    \annot
  \end{tikzpicture}
  \end{center}
  \vspace{0.1cm}
\end{minipage}
\hfill
\begin{minipage}[b]{0.18\linewidth}
  \begin{center}
  \begin{tikzpicture}[xscale=\imagescale,yscale=-\imagescale]
    \node[anchor=north west,inner sep=0pt,outer sep=0pt] at (-0.5,-0.5)
       {\includegraphics[width=\imagewidth]{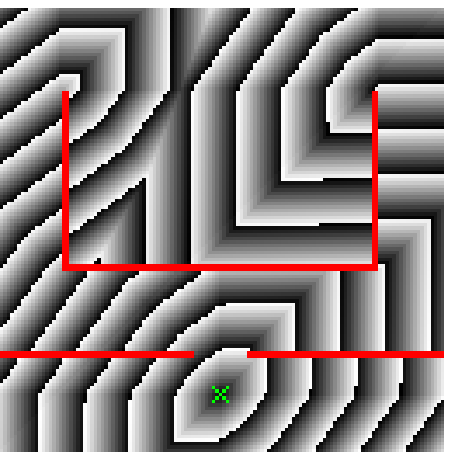}};
    \annot
  \end{tikzpicture}
  \end{center}
  \vspace{0.1cm}
\end{minipage}
\hfill
\begin{minipage}[b]{0.10\linewidth}
  \centerline{}
\end{minipage}
\begin{minipage}[b]{0.10\linewidth}
  \centerline{}
\end{minipage}
\hfill
\begin{minipage}[b]{0.18\linewidth}
  \begin{center}
  \begin{tikzpicture}[xscale=\imagescale,yscale=-\imagescale]
    \node[anchor=north west,inner sep=0pt,outer sep=0pt] at (-0.5,-0.5)
       {\includegraphics[width=\imagewidth]{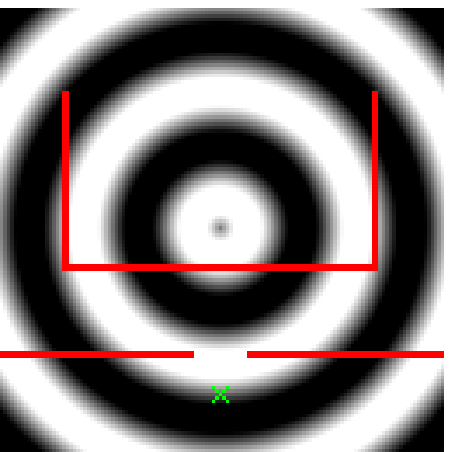}};
    \annot
  \end{tikzpicture}
  \end{center}
  \centerline{Geodesic mask}
  \vspace{0.2cm}
\end{minipage}
\hfill
\begin{minipage}[b]{0.18\linewidth}
  \begin{center}
  \begin{tikzpicture}[xscale=\imagescale,yscale=-\imagescale]
    \node[anchor=north west,inner sep=0pt,outer sep=0pt] at (-0.5,-0.5)
       {\includegraphics[width=\imagewidth]{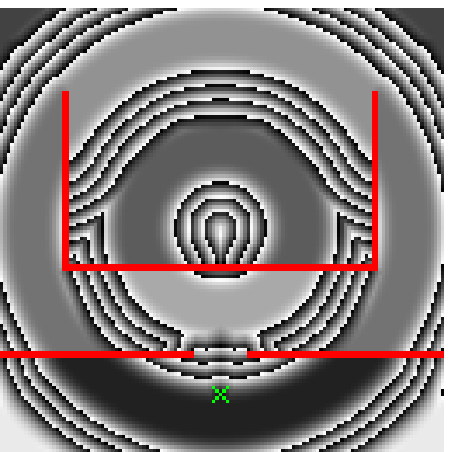}};
    \annot
  \end{tikzpicture}
  \end{center}
  \centerline{GRAYMAT}
  \vspace{0.2cm}
\end{minipage}
\hfill
\begin{minipage}[b]{0.18\linewidth}
  \begin{center}
  \begin{tikzpicture}[xscale=\imagescale,yscale=-\imagescale]
    \node[anchor=north west,inner sep=0pt,outer sep=0pt] at (-0.5,-0.5)
       {\includegraphics[width=\imagewidth]{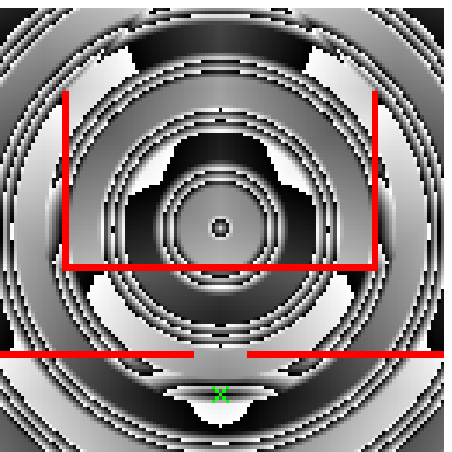}};
    \annot
  \end{tikzpicture}
  \end{center}
  \centerline{DOCS}
  \vspace{0.2cm}
\end{minipage}
\hfill
\begin{minipage}[b]{0.18\linewidth}
  \begin{center}
  \begin{tikzpicture}[xscale=\imagescale,yscale=-\imagescale]
    \node[anchor=north west,inner sep=0pt,outer sep=0pt] at (-0.5,-0.5)
       {\includegraphics[width=\imagewidth]{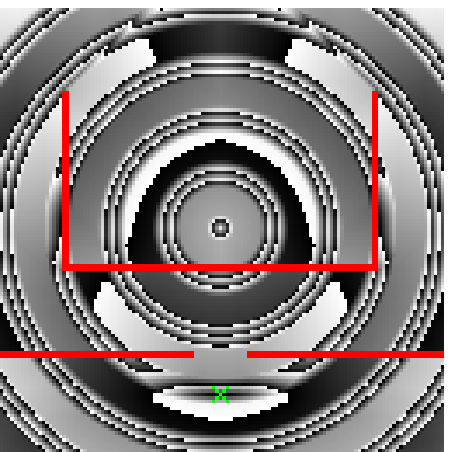}};
    \annot
  \end{tikzpicture}
  \end{center}
  \centerline{WDOCS}
  \vspace{0.2cm}
\end{minipage}
\hfill
\begin{minipage}[b]{0.10\linewidth}
  \centerline{}
\end{minipage}
\caption{Grey-weighted distance transforms calculated on two different geodesic masks using the 'optimal' chamfer weights. The left column shows the images used as geodesic masks: (top) a gradient image with value 0 in the top left corner and 255 in the bottom right corner; (bottom) a sinusoidal image with clamped amplitudes. The three remaining images on each row are the (from left to right) GRAYMAT, DOCS, and WDOCS transforms respectively. A cyclic grey-level palette has been used to visualise the geodesic fronts.}
\label{fig:transforms}
\end{figure*}

\figurename~\ref{fig:transforms} shows the different grey-weighted distance function behaviours. The top row shows the grey-weighted distance transforms when using a gradient image as geodesic mask. The GRAYMAT transform progress rapidly across the area of low grey levels in the top left corner while DOCS and WDOCS have their fastest progression when travelling normal to the gradient direction. The bottom row shows the transforms on a sinusoidal image with saturated intensities. Once again it is clear that GRAYMAT moves fast in areas of low grey level (the black rings) and slow in areas of high grey level (the white rings). DOCS and WDOCS, on the other hand, move fast in both areas with high and low grey level but move slower through the transitions between two uniform areas, where the difference in grey levels results in increased costs. A more detailed analysis of the behaviour of the different transforms is presented in~\cite{mg:fouardDGCI2006}.

\comment{
\begin{table}\label{tab:chamferweights}
\begin{center}
\caption{Length of elementary step for different distances. For 2D the weights are $<\mbox{edge neighbour},\mbox{vertex neighbour}>$, and for 3D the weights are $<\mbox{face neighbour},\mbox{edge neighbour},\mbox{vertex neighbour}>$.}
\medskip
\small{
\begin{tabular}{lcc}
\hline
\noalign{\smallskip}
& \multicolumn{2}{c}{Elementary step weights} \\
Distance & 2D & 3D \\
\hline
\noalign{\smallskip}
Chessboard \qquad \qquad & $<1,1>$ & $<1,1,1>$ \\
City block \qquad \qquad & $<1,2>$ & $<1,2,3>$ \\
Euclidean  & $<1,\sqrt{2}>$ & $<1,\sqrt{2},\sqrt{3}>$ \\
3-4-5 & $<3,4>$ & $<3,4,5>$ \\
Optimal & $<0.9604,1.3583>$ & $<0.9398,1.3291,1.6278>$ \\
\hline
\end{tabular}
}
\end{center}
\end{table}
}

\section{Algorithms}
\label{sec:algorithms}

When referring to a grey-weighted distance computation it can be in one of four settings: (i) {\it grey-weighted transform} (or {\it transform} for short), (ii) {\it seeded grey-weighted transform} (or {\it seeded transform} for short), (iii) {\it grey-weighted dilation} (or {\it dilation} for short), or (iv) {\it route}. In the seeded (or marker-based) distance transform, each pixel is the grey-weighted distance of the lowest cost path from a set $\mathcal{S}$ of predefined pixels generally referred to as {\it seeds}, {\it markers} or {\it features}. The other three settings, (i), (iii), and (iv) are all special cases of the seeded transform. In the distance transform, each pixel is the grey-weighted distance from the background. Generally the background is defined as $B = \{ x \, | \, f(x) = 0 \}$, {\it i.e.}, all pixels with grey level zero in the geodesic mask, which is the same as a seeded grey-weighted transform where $\mathcal{S} = \{x \, | \, f(x) = 0\}$. Dilation is a seeded transform where the seeds represent the region to be dilated and the calculation is stopped when a predefined grey-weighted distance is reached. The result is analogous to a morphological grey-level dilation of $\mathcal{S}$ where the structuring element is defined by the cost function. The route is a single source shortest path problem calculating the grey-weighted distance from a single pixel $p$ to a single pixel $q$. This is done by seeded transform from the single seed point $p$ and terminating the transform once $q$ is reached.

The seeded transform offers better options for experimental setups than the regular transform by facilitating multiple runs on the same image using different seeds. However,
the evaluation can be used as a guideline for choosing an algorithm for both the seeded and unseeded transform since the algorithms are the same. The runtime of a dilation or a route is naturally lower than that of calculating the transform of an entire image. However, the aim of this work is not to illustrate the low computational costs of various methods but to compare various implementations of grey-level--based distance computations, which is shown more clearly for complete (or near complete) image transforms than for dilations and routes.

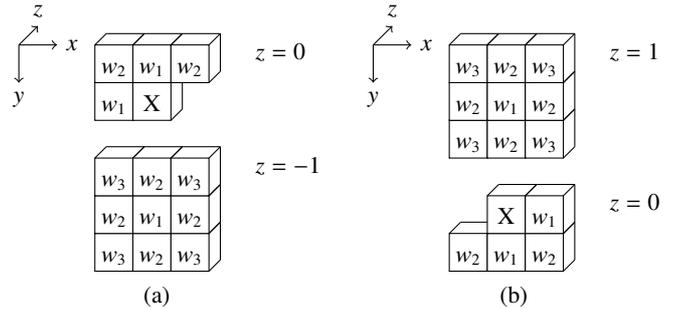
\begin{figure}
\newcommand{\drawVoxel}[3]{
  \draw (#1,#2) -- +(0.3,0.3);
  \draw (#1,#2)+(0,1) -- +(0.3,1.3);
  \draw (#1,#2)+(1,0) -- +(1.3,0.3);
  \draw (#1,#2)+(1,1) -- +(1.3,1.3);

  \draw (#1,#2)+(0.3,1.3) -- +(1.3,1.3);
  \draw[color=black,fill=white] (#1,#2) rectangle +(1,1);
  \node[xshift=0.25cm,yshift=0.2cm,anchor=mid] at (#1,#2) {#3};
}
\newcommand{\drawVoxelTopRight}[3]{
  \draw (#1,#2) -- +(0.3,0.3);
  \draw (#1,#2)+(0,1) -- +(0.3,1.3);
  \draw (#1,#2)+(1,0) -- +(1.3,0.3);
  \draw (#1,#2)+(1,1) -- +(1.3,1.3);

  \draw (#1,#2)+(1.3,0.3) -- +(1.3,1.3);
  \draw (#1,#2)+(0.3,1.3) -- +(1.3,1.3);
  \draw[color=black,fill=white] (#1,#2) rectangle +(1,1);
  \node[xshift=0.25cm,yshift=0.2cm,anchor=mid] at (#1,#2) {#3};
}
\newcommand{\drawVoxelRight}[3]{
  \draw (#1,#2) -- +(0.3,0.3);
  \draw (#1,#2)+(0,1) -- +(0.3,1.3);
  \draw (#1,#2)+(1,0) -- +(1.3,0.3);
  \draw (#1,#2)+(1,1) -- +(1.3,1.3);

  \draw (#1,#2)+(1.3,0.3) -- +(1.3,1.3);
  \draw[color=black,fill=white] (#1,#2) rectangle +(1,1);
  \node[xshift=0.25cm,yshift=0.2cm,anchor=mid] at (#1,#2) {#3};
}
\centering
\begin{minipage}[b]{0.45\columnwidth}
\centering
\small{
\begin{tikzpicture}[scale=0.5]
  \draw[->] (-2,9) -- (-1.5,9.5) node [above] {$z$};
  \draw[->] (-2,9) -- (-1,9) node [right] {$x$};
  \draw[->] (-2,9) -- (-2,8) node [below] {$y$};

  \drawVoxel{0}{3}{$w_3$};
  \drawVoxel{1}{3}{$w_2$};
  \drawVoxelRight{2}{3}{$w_3$};
  \drawVoxel{0}{4}{$w_2$};
  \drawVoxel{1}{4}{$w_1$};
  \drawVoxelRight{2}{4}{$w_2$};
  \drawVoxel{0}{5}{$w_3$};
  \drawVoxel{1}{5}{$w_2$};
  \drawVoxelTopRight{2}{5}{$w_3$};

  \drawVoxel{0}{7}{$w_1$};
  \drawVoxelRight{1}{7}{X};
  \drawVoxel{0}{8}{$w_2$};
  \drawVoxel{1}{8}{$w_1$};
  \drawVoxelTopRight{2}{8}{$w_2$};

  \node at (4,8.8) [right] {$z = 0$};
  \node at (4,5.8) [right] {$z = -1$};

\end{tikzpicture}
}
\centerline{(a)}
\end{minipage}
\hfill
\begin{minipage}[b]{0.49\columnwidth}
\centering
\small{
\begin{tikzpicture}[scale=0.5]
  \draw[->] (-2,9) -- (-1.5,9.5) node [above] {$z$};
  \draw[->] (-2,9) -- (-1,9) node [right] {$x$};
  \draw[->] (-2,9) -- (-2,8) node [below] {$y$};

  \drawVoxel{0}{6}{$w_3$};
  \drawVoxel{1}{6}{$w_2$};
  \drawVoxelRight{2}{6}{$w_3$};
  \drawVoxel{0}{7}{$w_2$};
  \drawVoxel{1}{7}{$w_1$};
  \drawVoxelRight{2}{7}{$w_2$};
  \drawVoxel{0}{8}{$w_3$};
  \drawVoxel{1}{8}{$w_2$};
  \drawVoxelTopRight{2}{8}{$w_3$};

  \node at (4,8.8) [right] {$z = 1$};

  \drawVoxel{0}{3}{$w_2$};
  \drawVoxel{1}{3}{$w_1$};
  \drawVoxelRight{2}{3}{$w_2$};
  \drawVoxel{1}{4}{X};
  \drawVoxelTopRight{2}{4}{$w_1$};

  \node at (4,4.8) [right] {$z = 0$};

\end{tikzpicture}
}
\centerline{(b)}
\end{minipage}
\caption{The masks for calculating grey-weighted distance using the chamfer algorithm for 3D images. The voxel position is marked with an X and $w_1$, $w_2$, and $w_3$ are the chamfer weights. (a) The mask used in the forward pass. (b) The mask used in the backward pass. The chamfer masks for 2D images are the masks above for $z = 0$.}
\label{fig:chamfermask}
\end{figure}

The first method to calculate grey-weighted distance transforms was to use the chamfer scan approach, {\it e.g.}, see~\cite{mg:GRAYMAT:LeviIC1970}. The algorithm uses a window containing a weight mask (chamfer mask), and is slided across the image, updating the central pixel at each position. The scan consists of a forward pass and a backward pass. Figure~\ref{fig:chamfermask} shows the masks used for the forward and backward passes for both 2D and 3D images. The chamfer weights are typically $w_1=3$, $w_2=4$, and $w_3=5$. In contrast with distances from binary images, the domain is usually not convex. Therefore, the chamfer algorithm for grey-weighted distance is an iterative process and has to be repeated until no updates are made in the distance map.

The chamfer algorithm can be considered an algorithm of the {\it label-correcting} kind. A grey-weighted distance label is assigned to a pixel at each step; the grey-weighted distance labels are estimates ({\it i.e.}, a upper bounds on) the grey-weighted distance of the lowest cost path from the source to the individual pixels. What characterises a label-correcting algorithm is that all labels are considered temporary until the final step, when they all become permanent.


A different approach from iterative raster scan in the chamfer algorithm is the graph search approach. Two simple graph search approaches are the depth-first search (DFS)~\cite{mg:Algorithms:Cormen1990} (referred to as {\it recursive propagation}) and the breadth-first search (BFS)~\cite{mg:Algorithms:Cormen1990} (referred to as {\it ordered propagation}), used by Silvela et al. for distance transform computations~\cite{mg:SilvelaITIP2001}. The algorithm for both approaches is listed in Algorithm~\ref{alg:propagation}. They are both label-correcting algorithms and the difference between them comes from how pixels are added and removed from the list $L$. For recursive propagation, $L$ is a last-in-first-out (LIFO) list, {\it i.e.}, a list where the last pixel added is the first to be removed. For ordered propagation, $L$ is a first-in-first-out (FIFO) list, {\it i.e.}, a list where the pixel added first is the first to be removed.

\begin{algorithm}[!ht]
  \caption{Depth/Breadth-first search (label-correcting)}\label{alg:propagation}
  \begin{algorithmic}[1]
    \REQUIRE Seed map $\mathcal{S}$, geodesic mask $f$, and empty list $L$.
    \ENSURE A grey-weighted distance map $G$ of $f$.
    \medskip
    \STATE set all elements of $G$ to $\infty$ except $\mathcal{S}$ which is set to $0$;
    \STATE put all pixels adjacent to $\mathcal{S}$, not on $\mathcal{S}$, in a list $L$;
    \WHILE{ $L$ is not empty }
      \STATE remove a pixel $x$ from $L$;
      \STATE find $d_{\text{min}} = \min_{n\in\text{adj}(x)}(G(x),G(n)+c(n,x))$;
      \IF{ $d_{min} < G(x)$ }
        \STATE set $G(x) = d_{min}$;
        \STATE put all pixels adjacent to $x$ on $L$;
      \ENDIF
    \ENDWHILE
  \end{algorithmic}
\end{algorithm}

The opposite of label-correcting algorithms are algorithms of the {\it label-setting} kind. A label-setting algorithm assigns one label as permanent (optimal) at each iteration. The algorithms of this group are basically various implementations of Dijkstra's well known algorithm~\cite{mg:DijkstraNM1959}, first proposed for grey-weighted distance transforms as the uniform cost algorithm~\cite{mg:UCADT:VerwerPAMI1989}. In graph search terminology it is referred to as best-first search since the best alternative is considered at every iteration. The label-setting algorithms are much more efficient than label-correcting algorithms, but they are applicable only to special situations like region growing scenarios. Recently much work on grey-weighted distance computations has been done using various label-setting implementations~\cite{mg:FalcaoTMI2000,mg:IkonenDGCI2005}. Even though all of them use the theoretically optimal Dijkstra's algorithm, they differ in what data structures they use. The label-setting algorithm used is listed in Algorithm~\ref{alg:bestfirst}.

\begin{algorithm}[!ht]
  \caption{Best-first (label-setting)}\label{alg:bestfirst}
  \begin{algorithmic}[1]
    \REQUIRE Seed map $\mathcal{S}$, geodesic mask $f$, empty priority queue $Q$, and empty set $E$ for expanded nodes.
    \ENSURE A grey-weighted distance map $G$ of $f$.
    \medskip
    \STATE set all elements of $G$ to $\infty$ except $\mathcal{S}$ which is set to $0$;
    \STATE put all pixels in $\mathcal{S}$ on the queue $Q$;
    \WHILE{ $Q$ is not empty }
      \STATE remove a pixel $x$ from $Q$ for which $G(x)$ is minimal;
      \STATE add $x$ to $E$;
      \FOR{ each $n$ adjacent to $x$ not in $E$}
        \STATE find $d_{\text{min}} = \min(G(n),G(x)+c(x,n))$;
        \IF{ $d_{min} < G(n)$ }
          \STATE set $G(n) = d_{min}$;
	  \IF{ $n$ is already on $Q$ }
            \STATE update position of $n$ in $Q$;
	  \ELSE
	    \STATE put $n$ on $Q$;
	  \ENDIF
        \ENDIF
      \ENDFOR
    \ENDWHILE
  \end{algorithmic}
\end{algorithm}

\subsection{Local cost computation}

The structure of the distance map will depend on the image connectivity and spatial distance between adjacent nodes. Common choices for spatial distance in a local neighbourhood are city block, chessboard, 3-4-5, or one of the optimal chamfer weights designed to approximate the Euclidean distance over large distances~\cite{mg:Chamfer:BorgeforsCVGIP1986,mg:BorgeforsCVIU1996,mg:VerwerPRL1991}. However, the relative efficiency of the various algorithms is unlikely to vary with different choices. We choose to only use the common 3-4-5 chamfer weights, since one of the algorithms is only applicable to integer costs, and, as previously mentioned, 8-connectivity for two-dimensional images and 26-connectivity for three-dimensional images.

\subsection{Implementations}

This section presents the details of the different strategies and data structures used for the test cases in Section~\ref{sec:experiments}. The methods are labelled using capital letters with subscripts representing the properties of the method/data structure (see \tablename~\ref{tab:methods} for a summary). These labels will be used throughout Section~\ref{sec:experiments}.


\subsubsection{Label-correcting algorithms}

The implementation of the chamfer algorithm does not give much room for variation. The chamfer method iterates over the image through raster scans using the chamfer mask shown in~\figurename~\ref{fig:chamfermask}, and is given the label C.

For recursive and ordered propagation we use variations of Algorithm~\ref{alg:propagation}. Step 8 puts all neighbours of $x$ on the list if $x$ is updated. This will result in lots of duplicates on the list and unnecessary pop operations. To improve the speed of the algorithm, a pointer array can be used to keep track of whether a pixel is already on the list. If a pixel is already on the list, it need not be duplicated. The propagation algorithms are given the label P with the subscript L for LIFO list (recursive propagation) and F for FIFO list (ordered propagation). The subscript A is used if a pointer array is used to keep track of which pixels are on the list.


\subsubsection{Label-setting algorithms using $d$-heap}

Roughness in images was computed by Ikonen et al.~\cite{mg:IkonenDGCI2005} using the DOCS transform with a binary heap~\cite{mg:Algorithms:Cormen1990} ($d$-ary heap with $d=2$). The algorithm used is the same as Algorithm~\ref{alg:bestfirst} but without Step 10, i.e., no check whether the neighbour $n$ is already on the queue $Q$. This results in duplicates on the queue, which leads to unnecessary pop operations. Here we represent the priority queue $Q$ in Algorithm~\ref{alg:bestfirst} by a $d$-heap, both without any helper structures, as Ikonen in~\cite{mg:IkonenDGCI2005}, and with some helper structures to keep track of the pixels on the queue. The key of a pixel $x$ in $Q$ is the grey-weighted distance $d(\mathcal{S},x)$ at the time it is inserted into $Q$. Since low grey-weighted distance values have priority over high values, all priority queues used in this work are minimum priority queues, {\it i.e.}, the root stores the element with the smallest key. Step 4 is the {\it extract-min} operation, which finds the smallest key and removes it from the heap, and the update in Step 11 is the {\it decrease-key} operation, which increases the node priority.

We use $d=2$ for all $d$-heaps in the experiments. See Section~\ref{sec:dheaps} for a more detailed discussion on selecting $d$. In the first implementation using $d$-heap,
we always insert a new instance of $n$ in Step 11 and 13, like in~\cite{mg:IkonenDGCI2005}, even if it means duplication. This algorithm is labelled H (for heap). In another implementation (labelled H\sub{A}) we use a pointer array, which for every pixel $x$ stores the position of $x$ in the heap or \texttt{NULL} indicating that $x$ is not on the heap. The pointer array is used in Step 10 to check if a pixel is on the queue, and in Step 11 to update the priority ({\it decrease-key}). The final group using $d$-heaps implement hash tables instead of a pointer array to use less memory. They use hash tables with various hash functions and various table sizes to keep track of the positions of the pixels in the heap. However, keeping a hash table requires additional computation, and hash tables work differently depending on the distribution of the key values assigned to the input data ({\it i.e.}, depending on the grey-weighted distance values and on the geometric structures of objects in the image).

We use the same geometric hash functions used by Nyul et al. for 3D images~\cite{mg:NyulGM2003}. They assign a key value to a heap pixel $x$ using the following equations:
\begin{eqnarray}
  \text{key}(x) &=& ((c_3 \cdot \text{height} + c_2) \cdot \text{width} + c_1) \, \text{modulo} \, H, \label{eq:hashlin} \\
  \text{key}(x) &=& (c_3 + c_2 + c_1) \, \text{modulo} \, H, \label{eq:hashsum} \\
  \text{key}(x) &=& (c_3 \cdot c_2 \cdot c_1) \, \text{modulo} \, H, \label{eq:hashprod} \\
  \text{key}(x) &=& (c_3 \, \oplus \, c_2 \, \oplus \, c_1) \, \text{modulo} \, H, \label{eq:hashxor}
\end{eqnarray}
where $c_1$, $c_2$, $c_3$ are the coordinates of the voxel, height and width are the dimensions of a slice, $H$ is the size of the hash table, and $\oplus$ is the bit-wise exclusive or operation. The corresponding hash functions for 2D are acquired by simply removing all instances of $c_3$ and 'height' from the above equations.


For Equations~(\ref{eq:hashlin}) and (\ref{eq:hashprod}) we use a hash table size of 512 in the 2D case and 8191 in the 3D case. The range of possible hash values is quite large and these sizes are reasonably small and result in a fairly uniform distribution of hash keys in our application. For Equation~(\ref{eq:hashsum}) we use a hash table size of 512 for the 2D case and 768 for the 3D case. For Equation~(\ref{eq:hashxor}) we use a size of 256. This gives us a separate hash bin for each combination of 8-bit coordinates. We use labels H\sub{LIN}, H\sub{SUM}, H\sub{PROD}, and H\sub{XOR}, respectively, for the algorithms corresponding to Equations~(\ref{eq:hashlin})-(\ref{eq:hashxor}).

\subsubsection{Label-setting algorithms using Fibonacci heap}

The Fibonacci heap~\cite{mg:Algorithms:Cormen1990} is a more complicated structure than the $d$-heap. While the $d$-heap supports in $O(\log n)$ worst-case time the operations {\it insert}, {\it extract-min}, {\it decrease-key} and {\it delete}, the Fibonacci heap supports the same operations but have the advantage that operations that do not involve deleting an element run in $O(1)$ amortised time~\cite{mg:Algorithms:Cormen1990}. This means that {\it insert} (Step 13) and {\it decrease-key} (Step 11) on average run in constant time. In theory the Fibonacci heap is especially desirable when the number of {\it extract-min} and {\it delete} is small relative to the number of other operations performed. However, the constant factors and programming complexity of Fibonacci heaps make them less desirable than ordinary $d$-heaps for most applications~\cite{mg:Algorithms:Cormen1990}. Despite the high programming complexity of the Fibonacci heap we have chosen to incorporate it for comparison purposes.

Like for $d$-heaps, the first version (labelled F for Fibonacci) does not keep track on whether a pixel is already on the heap, and we do not perform a search in the heap for a pixel already stored. This means that we always insert a new instance of the pixel, even if it means duplication. In another version (labelled F\sub{A}) we use a pointer array to keep track of pixels on the heap. The final version (labelled F\sub{SUM}) use the hash table that performed best among Equations~(\ref{eq:hashlin})-(\ref{eq:hashxor}) in Section~\ref{sec:2dtests} (Equation~(\ref{eq:hashsum})), with the same table size as was used for the $d$-heap.

\subsubsection{Label-setting algorithms using circular bucket queues}
\label{sec:bucketqueues}

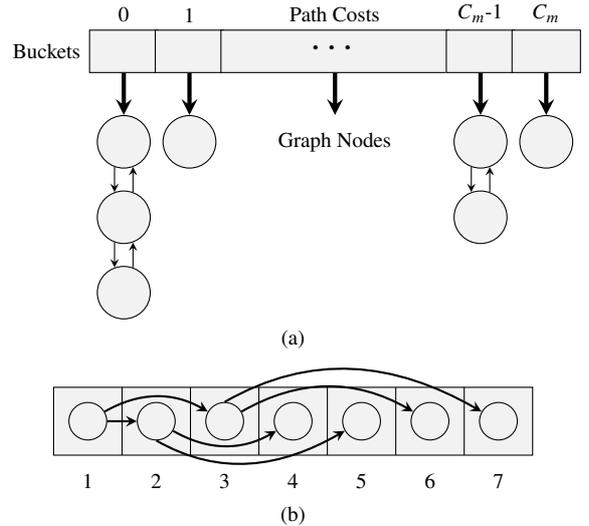
\begin{figure}
\tikzstyle{bucket}     = [rectangle,draw,fill=gray!10,anchor=west,inner sep=8pt,minimum width=0.9cm]
\tikzstyle{bigbucket}  = [style=bucket,minimum width=3cm]
\tikzstyle{node}       = [circle,draw,fill=gray!10,inner sep=7pt]
\tikzstyle{thickarrow} = [->,>=stealth,ultra thick]
\tikzstyle{thinarrow}  = [->,>=stealth]

\begin{minipage}[b]{\linewidth}
  \centering
  \footnotesize{
  \begin{tikzpicture}[xscale=1,yscale=1]
    

    \node[style=bucket] at (0,0) (b1) {};
    \node[style=bucket] at ($(b1.east)-(0.05,0)$) (b2) {};
    \node[style=bigbucket] at ($(b2.east)-(0.05,0)$) (b3) {};
    \node[style=bucket] at ($(b3.east)-(0.05,0)$) (b4) {};
    \node[style=bucket] at ($(b4.east)-(0.05,0)$) (b5) {};

    \node[left] at (b1.west) {Buckets};
    \node[above] at (b1.north) {0};
    \node[above] at (b2.north) {1};
    \node[above] at (b3.north) {Path Costs};
    \node[above] at (b4.north) {$C_m$-1};
    \node[above] at (b5.north) {$C_m$};
    \node at (b3) {\Large{$\cdots$}};

    \path (b1) ++(0,-1.2) node[style=node] (n11) {}
      ++(0,-1)  node[style=node] (n12) {}
      ++(0,-1)  node[style=node] (n13) {};
    \path (b2) ++(0,-1.2) node[style=node] (n21) {};
    \path (b3) ++(0,-1.2) node (n31) {Graph Nodes};
    \path (b4) ++(0,-1.2) node[style=node] (n41) {}
      ++(0,-1)  node[style=node] (n42) {};
    \path (b5) ++(0,-1.2) node[style=node] (n51) {};

    \draw[style=thickarrow] (b1.south) -- (n11.north);
    \draw[style=thickarrow] (b2.south) -- (n21.north);
    \draw[style=thickarrow] (b3.south) -- ($ (n21.north)!(b3.south)!(n41.north) $);
    \draw[style=thickarrow] (b4.south) -- (n41.north);
    \draw[style=thickarrow] (b5.south) -- (n51.north);

    \draw[style=thinarrow] (n11.250) -- (n12.110);
    \draw[style=thinarrow] (n12.70) -- (n11.290);
    \draw[style=thinarrow] (n12.250) -- (n13.110);
    \draw[style=thinarrow] (n13.70) -- (n12.290);

    \draw[style=thinarrow] (n41.250) -- (n42.110);
    \draw[style=thinarrow] (n42.70) -- (n41.290);
  \end{tikzpicture}
  }
  \centerline{(a)}
\end{minipage}
\begin{minipage}[b]{\linewidth}
  \centering
  \footnotesize{
  \begin{tikzpicture}[xscale=0.9,yscale=0.9]
    
    \path[draw=black,fill=gray!10] (0,0) rectangle (7,1);    
    \foreach \x in {1,2,3,4,5,6}
      \draw (\x,0) -- (\x,1);
    \foreach \x/\i/\name in {0.5/1/a,1.5/2/b,2.5/3/c,3.5/4/d,4.5/5/e,5.5/6/f,
                             6.5/7/g}
      \path (\x,0.5) node[circle,draw,fill=gray!10,inner sep=5pt] (\name) {}
      node[below=0.6cm] {$\i$};

    \path[->,>=stealth,thick] (a) edge (b);
    \path[->,>=stealth,thick] (a) edge[out=30,in=150] (c);
    \path[->,>=stealth,thick] (b) edge[out=-30,in=-150] (d);
    \path[->,>=stealth,thick] (b.south) edge[out=-30,in=-150] (e);
    \path[->,>=stealth,thick] (c) edge[out=30,in=150] (f);
    \path[->,>=stealth,thick] (c.north) edge[out=30,in=150] (g);

  \end{tikzpicture}
  }
  \centerline{(b)}
\end{minipage}
\caption{(a) Dial's priority queue based on buckets. The buckets are arranged linearly and sorted by path cost. Each bucket is associated with a list of graph nodes. (b) A small binary heap stored in an array.}
\label{fig:bucketqueue}
\end{figure}

Falcao et al. used an efficient implementation of Dijkstra's shortest path algorithm for grey-weighted distance~\cite{mg:FalcaoTMI2000}. It was the circular bucket queue data structure for integer costs introduced by Dial~\cite{mg:DialCACM1969}. The priority queue is represented by a circular array where every position (bucket) in the array holds a doubly linked list of all nodes with equal path cost. This is illustrated in~\figurename~\ref{fig:bucketqueue}~(a), where the priority queue is represented as an array of $B = C_m + 1$ buckets containing the nodes in $G$, with $C_m$ being the maximal possible arc weight. Each bucket $k$ stores a list of all nodes whose path cost is equal to $k$. The drawback with Dial's queue is that it only works with integer values, {\it i.e.}, in our case we can only run it with the DOCS cost function. We test Dial's bucket queue with both LIFO and FIFO lists, labelled D\sub{L} and D\sub{F} respectively. The LIFO version is also implemented using a pointer array to keep track of whether a pixel is on the queue or not. The pointer array implementation is labelled D\sub{LA}.


Yatziv et al. have proposed the Untidy queue~\cite{mg:YatzivJCP2006}. It is a circular bucket queue like Dial's, but the number of buckets can be chosen freely and each bucket stores all nodes whose path cost fall within a certain interval. If the number of buckets (referred to as {\it bucket size}) is $B = B_m + 1$, with $B_m$ being the highest bucket index, then the bucket number $k$ is determined by
\begin{equation}
  k = \text{floor}\left(\frac{d(p)}{C_m}B_m\right), \label{eq:bucket}
\end{equation}
where $d(p)$ is the path cost of node $p$ and $C_m$ is the maximal possible arc weight. It follows that each bucket $k$ will hold all nodes whose path cost is on the interval
\begin{equation}\label{eq:costinterval}
  d_k(p) = \left[k\frac{C_m}{B_m},\,\,  (k+1)\frac{C_m}{B_m} \right).
\end{equation}
This means that the Untidy priority queue works on floating point values as well as integer values. However, since the queue can put nodes with different costs in the same bucket, the Untidy queue will not necessarily return the node with the lowest path cost from a {\it find-min} operation. This introduces a rounding error which can be bound by choosing a large bucket size. See~\cite{mg:RaschJNA2009} for a more detailed analysis of the rounding error.
Here we choose the bucket size 261 for DOCS and 2551 for GRAYMAT and WDOCS to ensure a low error. See Section~\ref{sec:bucketsizes} for a more detailed discussion on selecting bucket sizes.
The labels for the Untidy queue implementations are U\sub{L}, U\sub{F}, and U\sub{LA}, where the subscripts refer to the same properties as for the Dial queue implementations.

The final circular bucket queue implementation used is the hierarchical heap proposed by Luengo~\cite{mg:LuengoIJCV2009}. The circular array functions the same way as for the Untidy queue, {\it i.e.}, each bucket holds pixels with costs on a given interval. However, instead of representing each bucket with a list the hierarchical heap uses a $d$-heap at each bucket. This introduces the extra runtime of the heap compared to a list, but it functions properly with real valued costs, i.e., it will always return the node with the lowest path cost from a {\it find-min} operation. The same bucket sizes were used as for the Untidy queue. The hierarchical heap was implemented in two versions. One without a pointer array to keep track of which pixels are on the queue (labelled HH) and one with a pointer array (labelled HH\sub{A}).

\tablename~\ref{tab:methods} summarises the various methods and their associated labels.

\begin{table}
\begin{center}
\caption{Summary of the various methods used for comparison and their associated labels}
\label{tab:methods}
\medskip
\footnotesize{
\begin{tabular}{llllll}
\hline
\noalign{\smallskip}
Label & Method & Data structure & P. array & Hashing \\
\hline
\noalign{\smallskip}
C           & Chamfer           &                &     &           \\
            &                   &                &     &           \\
P\sub{L}    & Depth-first       & LIFO           &     &           \\
P\sub{F}    & Breadth-first     & FIFO           &     &           \\
P\sub{LA}   & Depth-first       & LIFO           & Yes &           \\
P\sub{FA}   & Breadth-first     & FIFO           & Yes &           \\
            &                   &                &     &           \\
H           & Best-first        & d-ary heap     &     &           \\
H\sub{A}    & Best-first        & d-ary heap     & Yes &           \\
H\sub{LIN}  & Best-first        & d-ary heap     &     & Eq.~(\ref{eq:hashlin}) \\
H\sub{SUM}  & Best-first        & d-ary heap     &     & Eq.~(\ref{eq:hashsum}) \\
H\sub{PROD} & Best-first        & d-ary heap     &     & Eq.~(\ref{eq:hashprod}) \\
H\sub{XOR}  & Best-first        & d-ary heap     &     & Eq.~(\ref{eq:hashxor}) \\
            &                   &                &     &           \\
F           & Best-first        & Fibonacci heap &     &           \\
F\sub{A}    & Best-first        & Fibonacci heap & Yes &           \\
F\sub{SUM}  & Best-first        & Fibonacci heap &     & Eq.~(\ref{eq:hashsum}) \\
            &                   &                &     &           \\
D\sub{L}    & Best-first        & Dial's/LIFO    &     & G-w. dist. \\
D\sub{F}    & Best-first        & Dial's/FIFO    &     & G-w. dist. \\
D\sub{LA}   & Best-first        & Dial's/LIFO    & Yes & G-w. dist. \\
            &                   &                &     &           \\
U\sub{L}    & Best-first        & Untidy/LIFO    &     & G-w. dist. \\
U\sub{F}    & Best-first        & Untidy/FIFO    &     & G-w. dist. \\
U\sub{LA}   & Best-first        & Untidy/LIFO    & Yes & G-w. dist. \\
            &                   &                &     &           \\
HH          & Best-first        & H-heap         &     & G-w. dist. \\
HH\sub{A}   & Best-first        & H-heap         & Yes & G-w. dist. \\
\hline
\end{tabular}
}
\end{center}
\end{table}

\subsubsection{Technical details}\label{sec:technicaldetails}

For comparison and memory-conserving purposes, all data structures use dynamic containers. Each time a node is pushed on a queue the node is created on-the-fly and put on the queue represented by a dynamically allocated data structure. We have used containers from the C++ Standard Template Library (STL)~\cite{mg:Josuttis1999} where applicable to ensure a low programming complexity. This will make the algorithms easier to implement and, thus, encourage image analysts to adopt to the guidelines provided. The depth-first and breadth-first lists use the STL List container. The d-ary heap and hierarchical heap use the STL Vector container (the STL heap does not support the update in Step 11 of Algorithm~\ref{alg:bestfirst}). \figurename~\ref{fig:bucketqueue}~(b) illustrates how an array is used to store a heap. Dial's bucket queue and the Untidy bucket queue are both implemented using an STL List container for each bucket. When a pointer array is used to keep track of nodes on the queue, the queue needs to allow for random access to facilitate updates (Step 11 of Algorithm~\ref{alg:bestfirst}). Standard list containers do not support random access, which means that updating a node on the Dial or Untidy queue is associated with a search through the list occupied by the node. This search will most certainly result in higher runtime compared to using a customised list implementation allowing random access.

The execution time can be cut by implementing custom containers using static arrays or intelligent reallocation based on application-specific memory usage. However, such custom implementations are memory consuming and have high programming complexity, and algorithms based on such implementations are not easy to adopt. For example, implementing the Dial queue using a static array, which was done by Falcao et al.~\cite{mg:FalcaoTMI2000}, is not only nontrivial compared to using STL lists, but also memory consuming. Since the static array needs to be the same size as the image and contain two pointers per element, the static array will use at least twice the memory of the distance map. See~\cite{mg:CotoCG2007} for a more detailed discussion on the memory issue for static arrays in region growing algorithms. Because of both the high programming complexity and the large memory requirements, static array implementations are out of scope for this article. However, we include runtimes for static array implementations of the Dial queue and the Untidy queue in Section~\ref{sec:static} to give an indication of the tradeoff between dynamically allocated priority queues and priority queues using static arrays.

\section{Tests and results}
\label{sec:experiments}

In this section we do comparative tests of the algorithms described in Section~\ref{sec:algorithms} on both 2D and 3D data with varying image properties. All tests were carried out on an Intel Xeon CPU 3.60GHz 64-bit Dual core computer with 4 GB RAM running Red Hat Enterprise Linux 5 update 2. The algorithms were compiled with gcc v3.4 using the {\tt-O3} optimisation flag and no advantage was taken of the multiple core architecture of the CPU.

\subsection{Datasets}

\begin{figure*}
  \tikzstyle{pixel} = [inner sep=0.4pt,fill=black,circle]

  \newcommand{\annot}{%
    \node[pixel] at (33.2500,   16.2500) {};
    \node[pixel] at (33.2500,   56.2500) {};
    \node[pixel] at (33.2500,   96.2500) {};
    \node[pixel] at (33.2500,  136.2500) {};
    \node[pixel] at (33.2500,  176.2500) {};
    \node[pixel] at (33.2500,  216.2500) {};
    \node[pixel] at (33.2500,  256.2500) {};
    \node[pixel] at (63.2500,   38.2500) {};
    \node[pixel] at (63.2500,   78.2500) {};
    \node[pixel] at (63.2500,  118.2500) {};
    \node[pixel] at (63.2500,  158.2500) {};
    \node[pixel] at (63.2500,  198.2500) {};
    \node[pixel] at (63.2500,  238.2500) {};
    \node[pixel] at (63.2500,  278.2500) {};
    \node[pixel] at (93.2500,   16.2500) {};
    \node[pixel] at (93.2500,   56.2500) {};
    \node[pixel] at (93.2500,   96.2500) {};
    \node[pixel] at (93.2500,  136.2500) {};
    \node[pixel] at (93.2500,  176.2500) {};
    \node[pixel] at (93.2500,  216.2500) {};
    \node[pixel] at (93.2500,  256.2500) {};
    \node[pixel] at (123.2500,   38.2500) {};
    \node[pixel] at (123.2500,   78.2500) {};
    \node[pixel] at (123.2500,  118.2500) {};
    \node[pixel] at (123.2500,  158.2500) {};
    \node[pixel] at (123.2500,  198.2500) {};
    \node[pixel] at (123.2500,  238.2500) {};
    \node[pixel] at (123.2500,  278.2500) {};
    \node[pixel] at (153.2500,   16.2500) {};
    \node[pixel] at (153.2500,   56.2500) {};
    \node[pixel] at (153.2500,   96.2500) {};
    \node[pixel] at (153.2500,  136.2500) {};
    \node[pixel] at (153.2500,  176.2500) {};
    \node[pixel] at (153.2500,  216.2500) {};
    \node[pixel] at (153.2500,  256.2500) {};
    \node[pixel] at (183.2500,   38.2500) {};
    \node[pixel] at (183.2500,   78.2500) {};
    \node[pixel] at (183.2500,  118.2500) {};
    \node[pixel] at (183.2500,  158.2500) {};
    \node[pixel] at (183.2500,  198.2500) {};
    \node[pixel] at (183.2500,  238.2500) {};
    \node[pixel] at (183.2500,  278.2500) {};
    \node[pixel] at (213.2500,   16.2500) {};
    \node[pixel] at (213.2500,   56.2500) {};
    \node[pixel] at (213.2500,   96.2500) {};
    \node[pixel] at (213.2500,  136.2500) {};
    \node[pixel] at (213.2500,  176.2500) {};
    \node[pixel] at (213.2500,  216.2500) {};
    \node[pixel] at (213.2500,  256.2500) {};
   }

  \pgfmathsetlength{\imagewidth}{4.5cm} 
  \def\imagescale{0.01875}

\begin{minipage}[t]{0.06\linewidth}
  \centerline{}
\end{minipage}
\begin{minipage}[t]{0.27\linewidth}
  \begin{center}
  \begin{tikzpicture}[xscale=\imagescale,yscale=-\imagescale]
    \node[anchor=north west,inner sep=0pt,outer sep=0pt] at (-0.5,-0.5)
       {\includegraphics[width=\imagewidth]{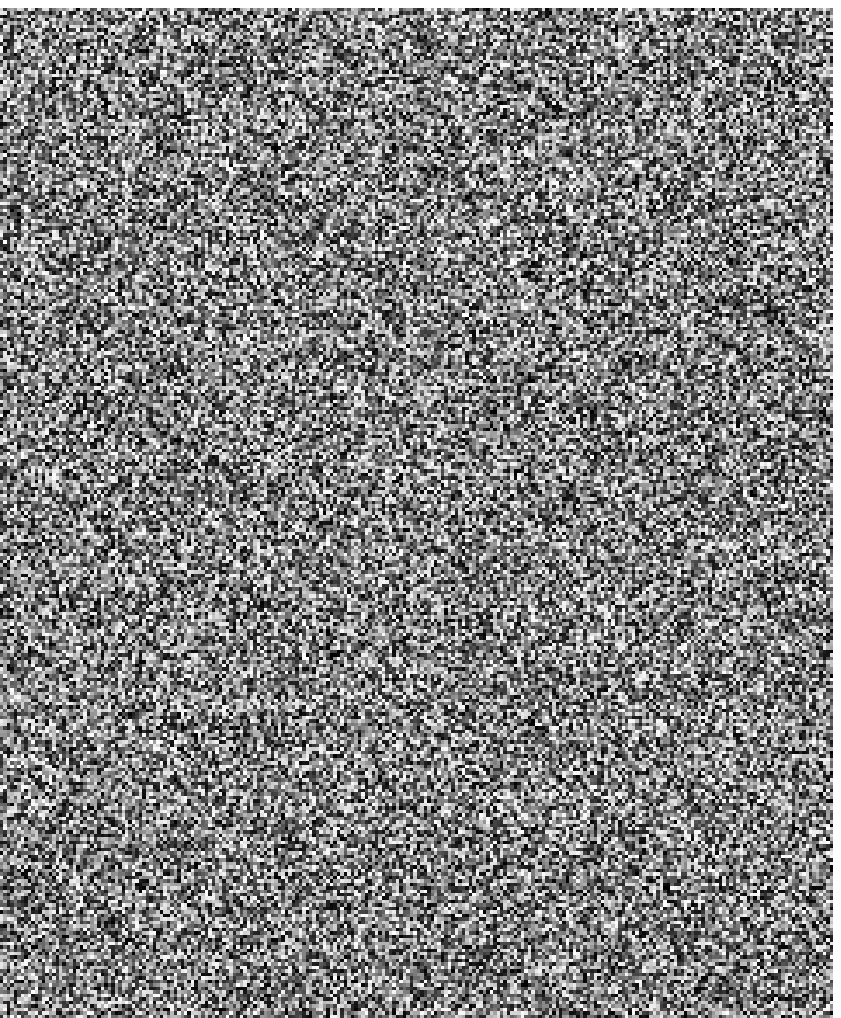}};
  \end{tikzpicture}
  \end{center}
  \centerline{}
\end{minipage}
\hfill
\begin{minipage}[t]{0.27\linewidth}
  \begin{center}
  \begin{tikzpicture}[xscale=\imagescale,yscale=-\imagescale]
    \node[anchor=north west,inner sep=0pt,outer sep=0pt] at (-0.5,-0.5)
       {\includegraphics[width=\imagewidth]{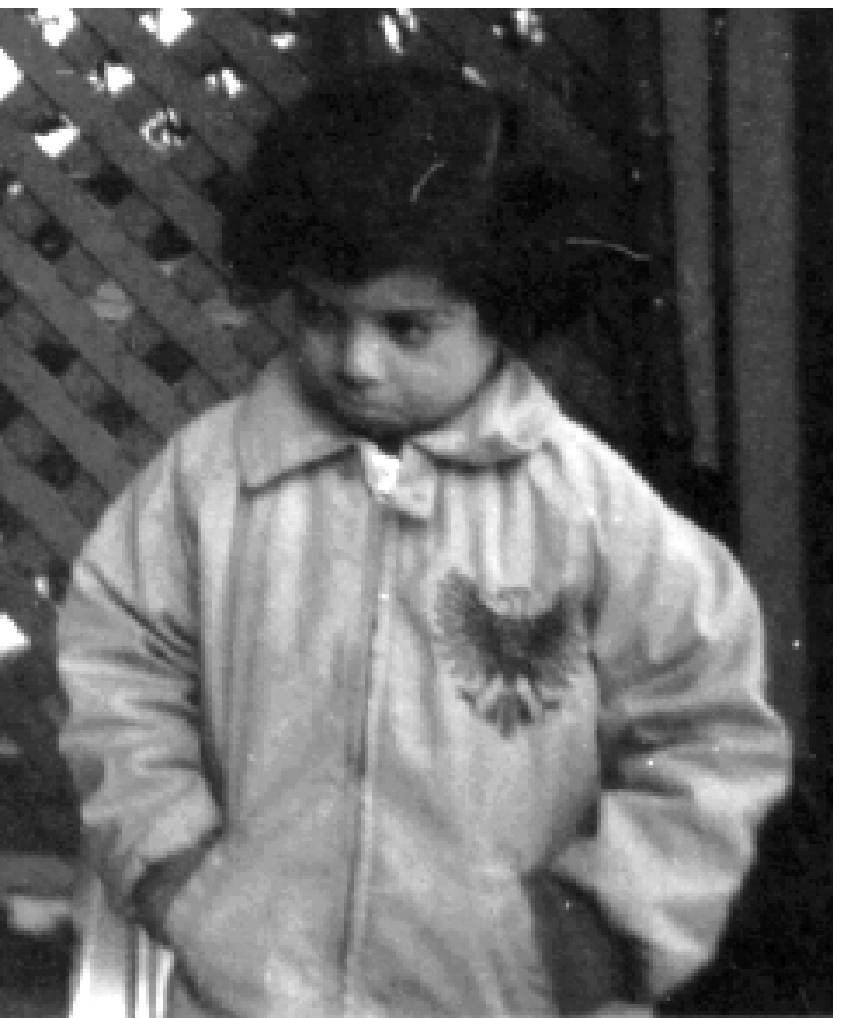}};
  \end{tikzpicture}
  \end{center}
  \centerline{}
\end{minipage}
\hfill
\begin{minipage}[t]{0.27\linewidth}
  \begin{center}
  \begin{tikzpicture}[xscale=\imagescale,yscale=-\imagescale]
    \node[anchor=north west,inner sep=0pt,outer sep=0pt] at (-0.5,-0.5)
       {\includegraphics[width=\imagewidth]{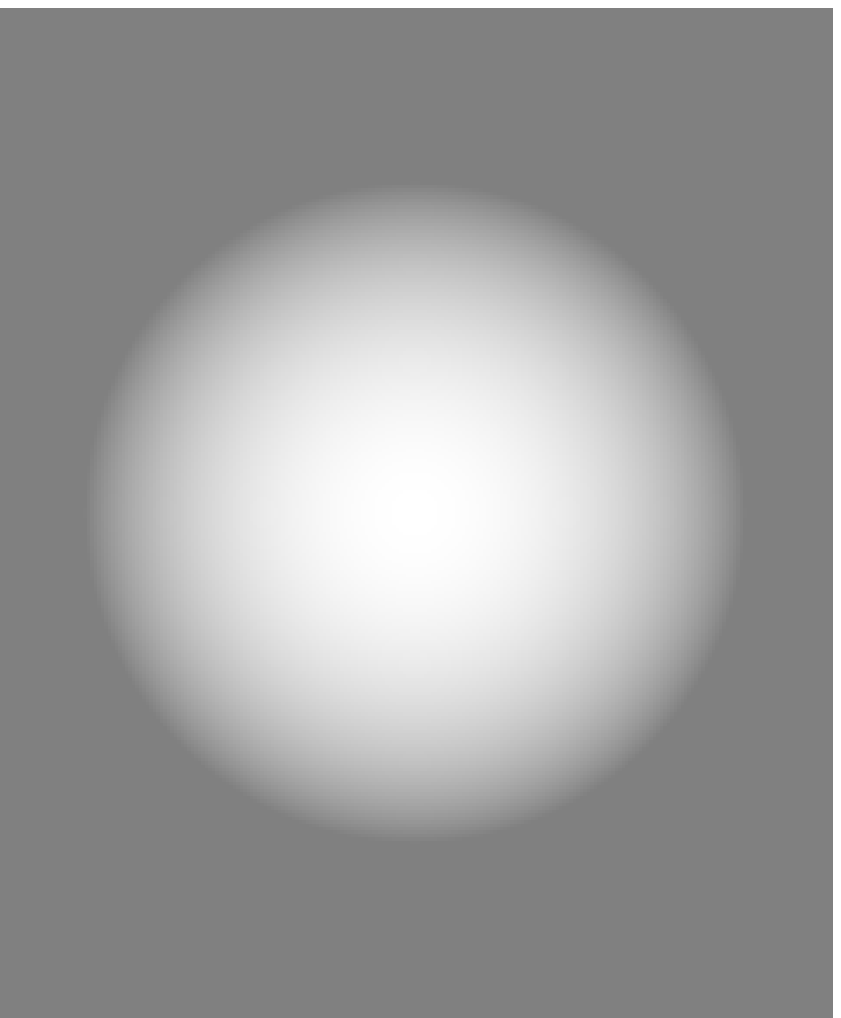}};
    \annot
  \end{tikzpicture}
  \end{center}
  \centerline{}
\end{minipage}
\begin{minipage}[t]{0.06\linewidth}
  \centerline{}
\end{minipage}
\begin{minipage}[t]{0.06\linewidth}
  \centerline{}
\end{minipage}
\begin{minipage}[b]{0.27\linewidth}
  \centering
  \begin{tikzpicture}[yscale=2,xscale=3]
    \begin{scope}
      \clip (-0.05,0) rectangle (1.05,1.1);
      \draw[very thin,color=gray!30,xstep=0.2,ystep=0.2] (0,0) grid +(1.1,1.1);
      \draw[color=blue!50!black,thick] plot file {hist_noise.table};
    \end{scope}
    \draw[->,>=stealth] (0,0) -- coordinate (x axis mid) (1.1,0);
    \draw[->,>=stealth] (0,0) -- coordinate (y axis mid) (0,1.2);
    \foreach \x/\xtext in {0.02/0,1/255}
    \draw[shift={(\x,0)}] (0pt,1pt) -- (0pt,-1pt) node[below] {\tick{$\xtext$}};
  \end{tikzpicture}
  \centerline{(a)}
\end{minipage}
\hfill
\begin{minipage}[b]{0.27\linewidth}
  \centering
  \begin{tikzpicture}[yscale=2,xscale=3]
    \begin{scope}
      \clip (-0.05,0) rectangle (1.05,1.1);
      \draw[very thin,color=gray!30,xstep=0.2,ystep=0.2] (0,0) grid +(1.1,1.1);
      \draw[color=blue!50!black,thick] plot file {hist_pout.table};
    \end{scope}
    \draw[->,>=stealth] (0,0) -- coordinate (x axis mid) (1.1,0);
    \draw[->,>=stealth] (0,0) -- coordinate (y axis mid) (0,1.2);
    \foreach \x/\xtext in {0.02/0,1/255}
    \draw[shift={(\x,0)}] (0pt,1pt) -- (0pt,-1pt) node[below] {\tick{$\xtext$}};
  \end{tikzpicture}
  \centerline{(b)}
\end{minipage}
\hfill
\begin{minipage}[b]{0.27\linewidth}
  \centering
  \begin{tikzpicture}[yscale=2,xscale=3]
    \begin{scope}
      \clip (-0.05,0) rectangle (1.05,1.1);
      \draw[very thin,color=gray!30,xstep=0.2,ystep=0.2] (0,0) grid +(1.1,1.1);
      \draw[color=blue!50!black,thick] plot file {hist_ball.table};
    \end{scope}
    \draw[->,>=stealth] (0,0) -- coordinate (x axis mid) (1.1,0);
    \draw[->,>=stealth] (0,0) -- coordinate (y axis mid) (0,1.2);
    \foreach \x/\xtext in {0.02/0,1/255}
    \draw[shift={(\x,0)}] (0pt,1pt) -- (0pt,-1pt) node[below] {\tick{$\xtext$}};
  \end{tikzpicture}
  \centerline{(c)}
\end{minipage}
\begin{minipage}[b]{0.06\linewidth}
  \centerline{}
\end{minipage}
\caption{Test images and their histograms for 2D transforms. (a) NOISE. A noisy image with uniform grey level. (b) POUT. A photograph. (c) BALL. A uniform image with the test point grid.}
\label{fig:testimages2d}
\end{figure*}

\begin{figure*}
\begin{minipage}[b]{0.05\linewidth}
  \centerline{}
\end{minipage}
\hfill
\begin{minipage}[b]{0.5\linewidth}
  \centerline{\includegraphics[width=\linewidth]{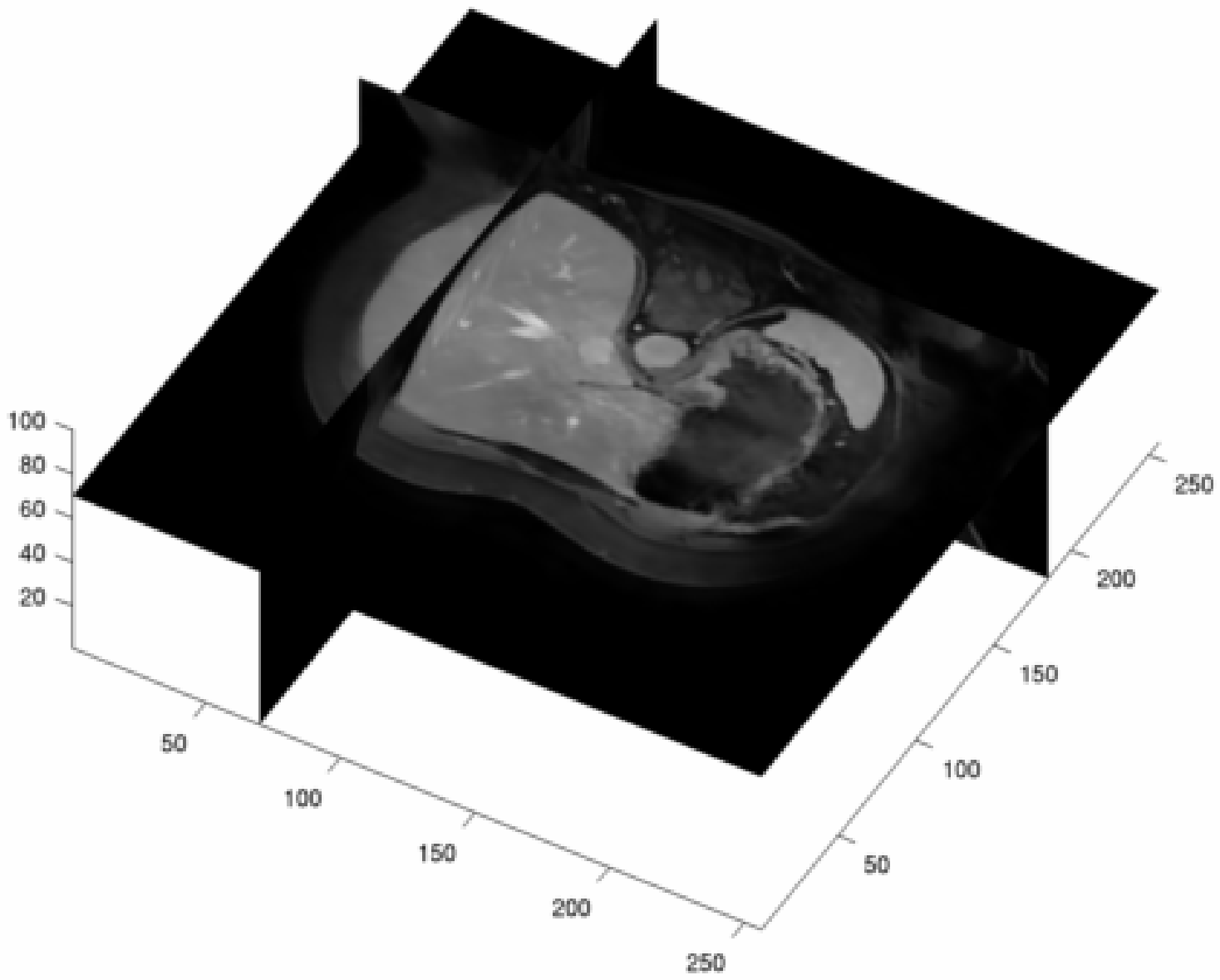}}
  \centerline{(a)}
\end{minipage}
\hfill
\begin{minipage}[b]{0.35\linewidth}
  \centerline{\includegraphics[width=\linewidth]{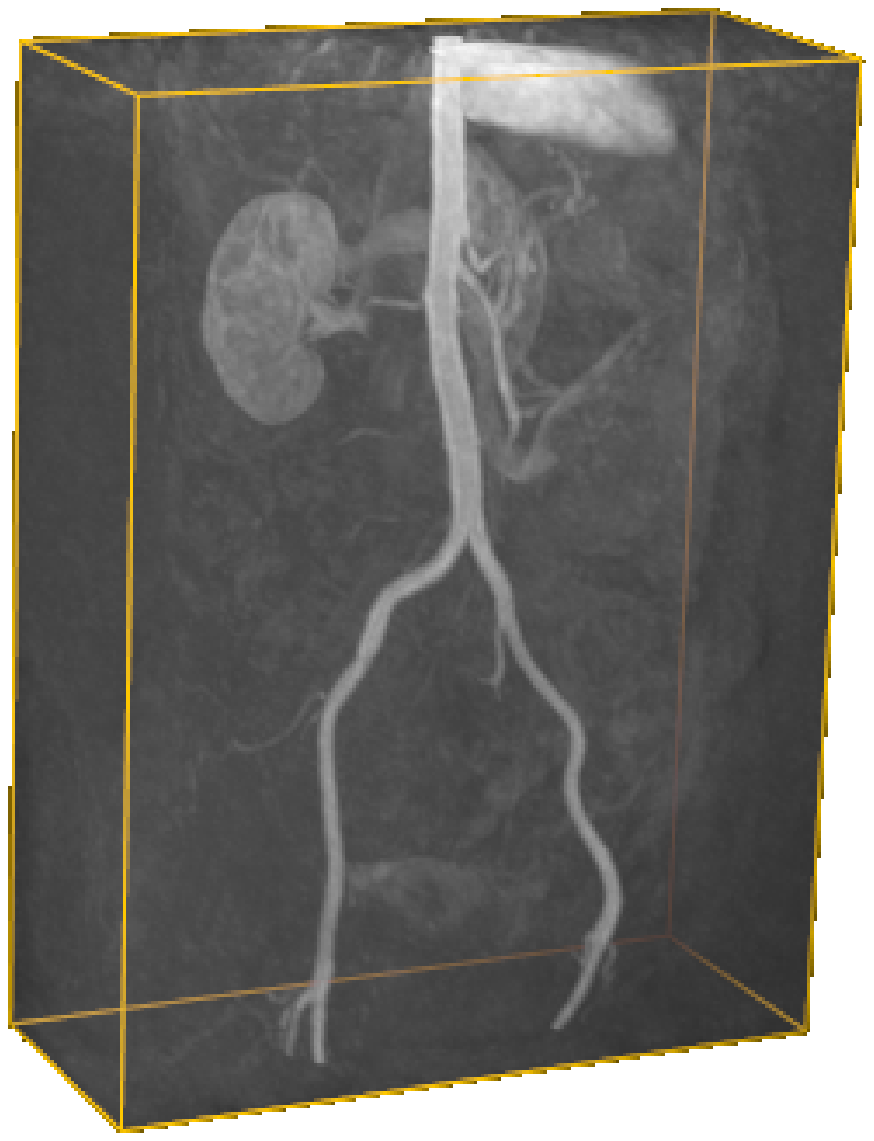}}
  \centerline{(b)}
\end{minipage}
\hfill
\begin{minipage}[b]{0.05\linewidth}
  \centerline{}
\end{minipage}
\caption{3D test images. (a) CT abdomen image of the liver region. (b) CE-MRA abdomen image of the aorta.}
\label{fig:testimages3d}
\end{figure*}

The algorithms were evaluated on images of various complexity to mimic the conditions common in image analysis problems. In the 2D case, the performance of the algorithms was compared using the 8-bit grey-level images seen in \figurename~\ref{fig:testimages2d}. The image in \figurename~\ref{fig:testimages2d}~(a) represents noisy images with high complexity and low spatial correlation between pixels; \figurename~\ref{fig:testimages2d}~(b) is the image {\tt pout.tif} from MATLAB\texttrademark (The MathWorks, Natick, MA, USA) and represents images of varying complexity with some spatial correlation; \figurename~\ref{fig:testimages2d}~(c) represents low complexity images with large uniform areas and, thus, high spatial correlation. Each image in \figurename~\ref{fig:testimages2d} was 960$\times$1164 pixels and are referred to as \mbox{NOISE}, \mbox{POUT}, and \mbox{BALL}, respectively. A test point grid was used for seeded grey-weighted distance transforms, see the black dots shown in \figurename~\ref{fig:testimages2d}~(c). The test point grid contain 49 points spread evenly, but not symmetrically, over the entire image area. In the 3D case, the performance of the algorithms was compared on an 8-bit gradient magnitude image of the 256$\times$256$\times$100 computed tomography (CT) image covering the liver region of an abdomen, shown in \figurename~\ref{fig:testimages3d}~(a), and a 512$\times$512$\times$154 contrast enhanced magnetic resonance angiography (CE-MRA) image of an abdomen, shown in \figurename~\ref{fig:testimages3d}~(b).

The CT image was used by Vidholm et al.~\cite{mg:VidholmMICCAI2006} to segment livers semi-automatically by seeded FMM. The CT images are abdominal contrast enhanced venous phase CT images of a patient with either carcinoid or endocrine pancreas tumour. The images were acquired with a Siemens Sensation 16 CT scanner. The CE-MRA image was used by Vidholm et al.~\cite{mg:VidholmISBI2004} for semi-automatic segmentation with haptic guided seeding. The image was acquired from a 1.5T Gyroscan Intera (Philips Medical Systems) using the standard body coil and a specially built table top extender. The sequence was a 3DRF-spoiled gradient echo with TR/TE/flip angle=$2.6/1.0/30^\circ$. The dataset consisted of four subvolumes: the head and upper thorax, the lower thorax and abdomen, the pelvis and upper legs, and the lower legs.

\subsection{Precomputed arc weights}

When calculating multiple transforms on the same image and the local cost function is complex, it might be favourable to precompute all arc weights to decrease the computational cost of each step. In these cases, precomputing arc weights can generally increase the efficiency at the cost of memory. If the arc weights are bi-directional, the look-up table of a 26-connected image will require roughly 13 times as much memory as the image. However, for grey-weighted distances the cost function is not very complex. Even if there is sufficient memory, the overhead from accessing the look-up table might be of the same magnitude as evaluating the cost function, in part because of cache misses. \tablename~\ref{tab:precomp} shows a sample of runtimes for two images when using both precomputed weights and when computing the weights on-the-fly.

\begin{table}
\begin{center}
\caption{Runtimes for using precomputed arc weights vs computing arc weights on-the-fly.}
\label{tab:precomp}
\medskip
\footnotesize{
\begin{tabular}{lllrrr}
\hline
\noalign{\smallskip}
Image & Alg.  & Definition & Pre. ($s$) & Transf. ($s$) & Total ($s$) \\
\hline
\noalign{\smallskip}
POUT     & H    & GRAYMAT &    - &  3.28 &  3.28 \\
         &      &         & 0.56 &  3.29 &  3.85 \\
         &      & WDOCS   &    - &  3.40 &  3.40 \\
         &      &         & 0.63 &  3.31 &  3.94 \\
         &      &         &      &       &       \\
CT       & H    & GRAYMAT &    - & 51.34 & 51.34 \\
         &      &         & 7.64 & 54.42 & 62.06 \\
         &      & WDOCS   &    - & 86.81 & 86.81 \\
         &      &         & 8.83 & 90.90 & 99.73 \\
\hline
\end{tabular}
}
\end{center}
\end{table}

In \tablename~\ref{tab:precomp} we see that the total runtime is higher for all cases where precomputed arc weights are used. For the three-dimensional case, the transform itself takes even longer to compute when using precomputed arc weights. There is apparently an overhead associated with using a (large) lookup table for the arc weight which is greater than computing the arc weight on-the-fly. Since the lookup table is memory consuming and the numbers in \tablename~\ref{tab:precomp} indicate that there is no gain in computation speed, we do not recommend precomputing arc weights and all experiments in Section~\ref{sec:experiments} calculate the arc weights on-the-fly.

\subsection{Different $d$-heaps}\label{sec:dheaps}

\tablename~\ref{tab:dheap} lists the runtimes for $d$-heaps with different $d$. It is apparent that the binary heap ($d=2$) performs best. It should be noted that the implementation differs slightly for the two cases $d=2$ and $d>2$. For the binary case ($d=2$), each element \texttt{a[i]} has children \texttt{a[2i]}, \texttt{a[2i+1]}, and parent \texttt{a[floor(i/2)]} (the root is \texttt{a[1]} and tree elements are \texttt{a[1]} $\ldots$ \texttt{a[n]}). See \figurename~\ref{fig:bucketqueue}~(b) for an illustration. For general $d$-heaps the child nodes are \texttt{a[d*(i-1)+2]} through \texttt{a[d*(i)+1]}, and the parent \texttt{a[floor((i-2)/d)+1]}. Because the $d$-heap gets shallower with increasing $d$ the {\it insert} operation will be faster but the {\it delete-min} operation will be more expensive due to the need of $d-1$ comparisons to find the smallest child node. It appears that for grey-weighted distance transforms, the slower {\it delete-min} operations, in combination with the extra multiplications and divisions by $d$ for the parent/child index calculation, makes the binary heap the preferred choice over any $d>2$.

\begin{table}
\begin{center}
\caption{Runtimes of $d$-heaps with different $d$.}
\label{tab:dheap}
\medskip
\footnotesize{
\begin{tabular}{ll*{6}{r}}
\hline
\noalign{\smallskip}
Dataset  & Alg. & \multicolumn{6}{c}{Runtime $(s)$} \\
\cline{3-8}
\noalign{\smallskip}
 & & $d=2$  & $d=3$  & $d=4$  & $d=5$  & $d=6$ & $d=7$ \\
\hline
\noalign{\smallskip}
NOISE    & H          & 3.59 & 3.76 & 3.75 & 3.77 & 3.78 & 3.84 \\
         & H\sub{A}   & 2.55 & 2.67 & 2.66 & 2.65 & 2.67 & 2.68 \\
         & H\sub{SUM} & 4.08 & 4.34 & 4.31 & 4.19 & 4.20 & 4.21 \\
& &  & & & & \\
POUT     & H           & 2.58 & 2.76 & 2.76 & 2.77 & 2.80 & 2.81 \\
         & H\sub{A}    & 2.35 & 2.47 & 2.47 & 2.48 & 2.49 & 2.51 \\
         & H\sub{SUM}  & 3.51 & 3.65 & 3.63 & 3.65 & 3.66 & 3.68 \\
& &  & & & & \\
BALL     & H           & 2.10 & 2.26 & 2.25 & 2.27 & 2.29 & 2.31 \\
         & H\sub{A}    & 2.27 & 2.39 & 2.39 & 2.41 & 2.42 & 2.45 \\
         & H\sub{SUM}  & 3.16 & 3.31 & 3.28 & 3.31 & 3.32 & 3.32 \\
\hline
\end{tabular}
}
\end{center}
\end{table}

\comment{
\begin{table}
\begin{center}
\caption{Runtimes of $d$-heaps with different $d$.}
\label{tab:dheap}
\medskip
\footnotesize{
\begin{tabular}{ll*{8}{r}}
\hline
\noalign{\smallskip}
Dataset  & Alg. & \multicolumn{8}{c}{Runtime $(s)$} \\
\cline{3-10}
\noalign{\smallskip}
 & & $d=2$  & $d=3$  & $d=4$  & $d=5$  & $d=6$ & $d=7$  & $d=8$  & $d=9$ \\
\hline
\noalign{\smallskip}
NOISE & 2000 & 3.59 & 3.76 & 3.75 & 3.77 & 3.78 & 3.84 & 3.86 & 3.89 \\
          & 2010 & 2.55 & 2.67 & 2.66 & 2.65 & 2.67 & 2.68 & 2.70 & 2.73 \\
          & 2021 & 4.08 & 4.34 & 4.31 & 4.19 & 4.20 & 4.21 & 4.23 & 4.27 \\
& &  & & & & \\
POUT & 2000 & 2.58 & 2.76 & 2.76 & 2.77 & 2.80 & 2.81 & 2.85 & 2.89 \\
         & 2010 & 2.35 & 2.47 & 2.47 & 2.48 & 2.49 & 2.51 & 2.57 & 2.63 \\
         & 2021 & 3.51 & 3.65 & 3.63 & 3.65 & 3.66 & 3.68 & 3.71 & 3.74 \\
& &  & & & & \\
BALL & 2000 & 2.10 & 2.26 & 2.25 & 2.27 & 2.29 & 2.31 & 2.34 & 2.36 \\
         & 2010 & 2.27 & 2.39 & 2.39 & 2.41 & 2.42 & 2.45 & 2.49 & 2.56 \\
         & 2021 & 3.16 & 3.31 & 3.28 & 3.31 & 3.32 & 3.32 & 3.37 & 3.38 \\
\hline
\end{tabular}
}
\end{center}
\end{table}
}

\subsection{Bucket sizes}\label{sec:bucketsizes}

Bucket sizes should be chosen so that a good tradeoff is reached between the number of buckets and the number of pixels in each bucket. Dial's queue defaults to bucket size $B = C_m + 1$ as mentioned in Section~\ref{sec:bucketqueues}. This assigns one bucket to each possible path cost. For the Untidy queue and the hierarchical heap, which both can handle floating point path costs, the right amount of buckets comes from a combination of grey-weighted distance definition, chamfer weight and actual runtime. To ensure a unique bucket for each possible path cost, we have from Equation~(\ref{eq:costinterval}) that the bucket size $B = B_m + 1$ must be chosen as,
\[
  B = \frac{C_m}{\min\left(\Delta d\right)} + 1,
\]
where $C_m$ is the maximum possible arc weight and $\min(\Delta d)$ is the minimum possible difference in path cost. \tablename~\ref{tab:bucketsize} lists the required number of buckets for each grey-weighted distance definition when using 8-bit images and 3-4-5 chamfer weights.

\begin{table}
\begin{center}
\caption{Properties of grey-weighted distance definitions for calculating bucket sizes when using 8-bit images and 3-4-5 chamfer weights.}
\label{tab:bucketsize}
\medskip
\footnotesize{
\begin{tabular}{lllllll}
\hline
\noalign{\smallskip}
 & & GRAYMAT & DOCS & WDOCS \\
\hline
\noalign{\smallskip}
Max cost                & $C_m$ & 1275 & 260 & $\sim$255.05 \\
Min path cost diff      & $\min(\Delta d)$ & 0.5 & 1 & $\sim$0.0137 \\
Buckets (unique cost)   & $B$ & 2551 & 261 & $>$18000 \\
\hline
\end{tabular}
}
\end{center}
\end{table}

\subsubsection{Untidy queue}

\begin{figure*}
  \def\mpscale{0.325}

  \begin{minipage}[t]{\mpscale\linewidth}
  \def\yscale{1.1}
  \def\xscale{0.015}
  \def\xmin{0}
  \def\xmax{310}
  \def\gridx{310}
  \def\ytitle{0.5}
  \def\uymin{1.75}
  \def\uymax{4.60}
  \def\ugridy{2.85}
  \def\myx{1/1,100/100,200/200,300/300}

  \centering
  \begin{tikzpicture}[yscale=\yscale,xscale=\xscale]
    \draw[very thin,color=gray!30,xstep=50,ystep=0.5] (\xmin,\uymin)
    grid +(\gridx,\ugridy);
      
    \draw[nd] plot file {tab_bs_rt_2300_09_110_rt.table};
    \draw[pd] plot file {tab_bs_rt_2300_10_110_rt.table};
    \draw[bd] plot file {tab_bs_rt_2300_11_110_rt.table};

    \draw[->,>=stealth] (\xmin,\uymin) --
    coordinate (x axis mid) (\xmax,\uymin);
    \draw[->,>=stealth] (\xmin,\uymin) --
    coordinate (y axis mid) (\xmin,\uymax) node[above] {$t(s)$};

    \path (x axis mid) +(0,\uymax-\uymin+\ytitle) node {DOCS};

    \foreach \x/\xtext in \myx
    \draw[shift={(\x,\uymin)}] (0pt,1pt) -- (0pt,-1pt)
    node[below] {\tick{$\xtext$}};

    \foreach \y in {2.0, 3.0, 4.0}
    \draw[shift={(\xmin,\y)}] (100pt,0pt) -- (-100pt,0pt)
    node[left] {\tick{$\y$}};
  \end{tikzpicture}
  \centerline{}
  \end{minipage}
  \hfill
  \begin{minipage}[t]{\mpscale\linewidth}
  \def\yscale{1.1}
  \def\xscale{0.015}
  \def\xmin{0}
  \def\xmax{310}
  \def\gridx{310}
  \def\ytitle{0.5}
  \def\uymin{1.75}
  \def\uymax{4.60}
  \def\ugridy{2.85}
  \def\myx{1/1,100/1000,200/2000,300/3000}

  \centering
  \begin{tikzpicture}[yscale=\yscale,xscale=\xscale]
    \draw[very thin,color=gray!30,xstep=50,ystep=0.5] (\xmin,\uymin)
    grid +(\gridx,\ugridy);
      
    \draw[ng] plot file {tab_bs_rt_2300_09_410_rt.table};
    \draw[pg] plot file {tab_bs_rt_2300_10_410_rt.table};
    \draw[bg] plot file {tab_bs_rt_2300_11_410_rt.table};

    \draw[->,>=stealth] (\xmin,\uymin) --
    coordinate (x axis mid) (\xmax,\uymin);
    \draw[->,>=stealth] (\xmin,\uymin) --
    coordinate (y axis mid) (\xmin,\uymax) node[above] {$t(s)$};

    \path (x axis mid) +(0,\uymax-\uymin+\ytitle) node {GRAYMAT};

    \foreach \x/\xtext in \myx
    \draw[shift={(\x,\uymin)}] (0pt,1pt) -- (0pt,-1pt)
    node[below] {\tick{$\xtext$}};

    \foreach \y in {2.0, 3.0, 4.0}
    \draw[shift={(\xmin,\y)}] (100pt,0pt) -- (-100pt,0pt)
    node[left] {\tick{$\y$}};
  \end{tikzpicture}
  \centerline{}
  \end{minipage}
  \hfill
  \begin{minipage}[t]{\mpscale\linewidth}
  \def\yscale{1.1}
  \def\xscale{0.023}
  \def\xmin{0}
  \def\xmax{170}
  \def\gridx{170}
  \def\ytitle{0.5}
  \def\uymin{1.75}
  \def\uymax{4.60}
  \def\ugridy{2.85}
  \def\myx{0.01/1,40/4000,80/8000,120/12000,160/16000}

  \centering
  \begin{tikzpicture}[yscale=\yscale,xscale=\xscale]
    \draw[very thin,color=gray!30,xstep=20,ystep=0.5] (\xmin,\uymin)
    grid +(\gridx,\ugridy);

    \draw[nw] plot file {tab_bs_rt_2300_09_210_rt.table};
    \draw[pw] plot file {tab_bs_rt_2300_10_210_rt.table};
    \draw[bw] plot file {tab_bs_rt_2300_11_210_rt.table};
      
    \draw[->,>=stealth] (\xmin,\uymin) --
    coordinate (x axis mid) (\xmax,\uymin);
    \draw[->,>=stealth] (\xmin,\uymin) --
    coordinate (y axis mid) (\xmin,\uymax) node[above] {$t(s)$};

    \path (x axis mid) +(0,\uymax-\uymin+\ytitle) node {WDOCS};

    \foreach \x/\xtext in \myx
    \draw[shift={(\x,\uymin)}] (0pt,1pt) -- (0pt,-1pt)
    node[below] {\tick{$\xtext$}};

    \foreach \y in {2.0, 3.0, 4.0}
    \draw[shift={(\xmin,\y)}] (70pt,0pt) -- (-70pt,0pt)
    node[left] {\tick{$\y$}};
  \end{tikzpicture}
  \centerline{}
  \end{minipage}
  \begin{minipage}[t]{\mpscale\linewidth}
  \def\yscale{1.1}
  \def\xscale{0.015}
  \def\xmin{0}
  \def\xmax{310}
  \def\gridx{310}
  \def\ytitle{0.5}
  \def\uymin{1.75}
  \def\uymax{4.60}
  \def\ugridy{2.85}
  \def\myx{1/1,100/100,200/200,300/300}

  \centering
  \begin{tikzpicture}[yscale=\yscale,xscale=\xscale]
    \draw[very thin,color=gray!30,xstep=50,ystep=0.5] (\xmin,\uymin)
    grid +(\gridx,\ugridy);

    \draw[nd] plot file {tab_bs_rt_2301_09_110_rt.table};
    \draw[pd] plot file {tab_bs_rt_2301_10_110_rt.table};
    \draw[bd] plot file {tab_bs_rt_2301_11_110_rt.table};
      
    \draw[->,>=stealth] (\xmin,\uymin) --
    coordinate (x axis mid) (\xmax,\uymin);
    \draw[->,>=stealth] (\xmin,\uymin) --
    coordinate (y axis mid) (\xmin,\uymax) node[above] {$t(s)$};

    \node[below=0.7cm] at (x axis mid) {no. of buckets};

    \foreach \x/\xtext in \myx
    \draw[shift={(\x,\uymin)}] (0pt,1pt) -- (0pt,-1pt)
    node[below] {\tick{$\xtext$}};

    \foreach \y in {2.0, 3.0, 4.0}
    \draw[shift={(\xmin,\y)}] (100pt,0pt) -- (-100pt,0pt)
    node[left] {\tick{$\y$}};
  \end{tikzpicture}
  \centerline{}
  \end{minipage}
  \hfill
  \begin{minipage}[t]{\mpscale\linewidth}
  \def\yscale{1.1}
  \def\xscale{0.015}
  \def\xmin{0}
  \def\xmax{310}
  \def\gridx{310}
  \def\ytitle{0.5}
  \def\uymin{1.75}
  \def\uymax{4.60}
  \def\ugridy{2.85}
  \def\myx{1/1,100/1000,200/2000,300/3000}

  \centering
  \begin{tikzpicture}[yscale=\yscale,xscale=\xscale]
    \draw[very thin,color=gray!30,xstep=50,ystep=0.5] (\xmin,\uymin)
    grid +(\gridx,\ugridy);

    \draw[ng] plot file {tab_bs_rt_2301_09_410_rt.table};
    \draw[pg] plot file {tab_bs_rt_2301_10_410_rt.table};
    \draw[bg] plot file {tab_bs_rt_2301_11_410_rt.table};
      
    \draw[->,>=stealth] (\xmin,\uymin) --
    coordinate (x axis mid) (\xmax,\uymin);
    \draw[->,>=stealth] (\xmin,\uymin) --
    coordinate (y axis mid) (\xmin,\uymax) node[above] {$t(s)$};

    \node[below=0.7cm] at (x axis mid) {no. of buckets};

    \foreach \x/\xtext in \myx
    \draw[shift={(\x,\uymin)}] (0pt,1pt) -- (0pt,-1pt)
    node[below] {\tick{$\xtext$}};

    \foreach \y in {2.0, 3.0, 4.0}
    \draw[shift={(\xmin,\y)}] (100pt,0pt) -- (-100pt,0pt)
    node[left] {\tick{$\y$}};
  \end{tikzpicture}
  \centerline{}
  \end{minipage}
  \hfill
  \begin{minipage}[t]{\mpscale\linewidth}
  \def\yscale{1.1}
  \def\xscale{0.023}
  \def\xmin{0}
  \def\xmax{170}
  \def\gridx{170}
  \def\ytitle{0.5}
  \def\uymin{1.75}
  \def\uymax{4.60}
  \def\ugridy{2.85}
  \def\myx{0.01/1,40/4000,80/8000,120/12000,160/16000}

  \centering
  \begin{tikzpicture}[yscale=\yscale,xscale=\xscale]
    \draw[very thin,color=gray!30,xstep=20,ystep=0.5] (\xmin,\uymin)
    grid +(\gridx,\ugridy);

    \draw[nw] plot file {tab_bs_rt_2301_09_210_rt.table};
    \draw[pw] plot file {tab_bs_rt_2301_10_210_rt.table};
    \draw[bw] plot file {tab_bs_rt_2301_11_210_rt.table};
      
    \draw[->,>=stealth] (\xmin,\uymin) --
    coordinate (x axis mid) (\xmax,\uymin);
    \draw[->,>=stealth] (\xmin,\uymin) --
    coordinate (y axis mid) (\xmin,\uymax) node[above] {$t(s)$};

    \node[below=0.7cm] at (x axis mid) {no. of buckets};

    \foreach \x/\xtext in \myx
    \draw[shift={(\x,\uymin)}] (0pt,1pt) -- (0pt,-1pt)
    node[below] {\tick{$\xtext$}};

    \foreach \y in {2.0, 3.0, 4.0}
    \draw[shift={(\xmin,\y)}] (70pt,0pt) -- (-70pt,0pt)
    node[left] {\tick{$\y$}};
  \end{tikzpicture}
  \centerline{}
  \end{minipage}
  \begin{minipage}[t]{\linewidth}
  \centering
  \plotlegend
  \end{minipage}
\caption{Runtimes for grey-weighted distance transforms when using different bucket sizes for the Untidy LIFO queue (top) and FIFO queue (bottom). Note the different scales on the x-axis for the different grey-weighted distance definitions.}
\label{fig:bucketsizesuntidy}
\end{figure*}
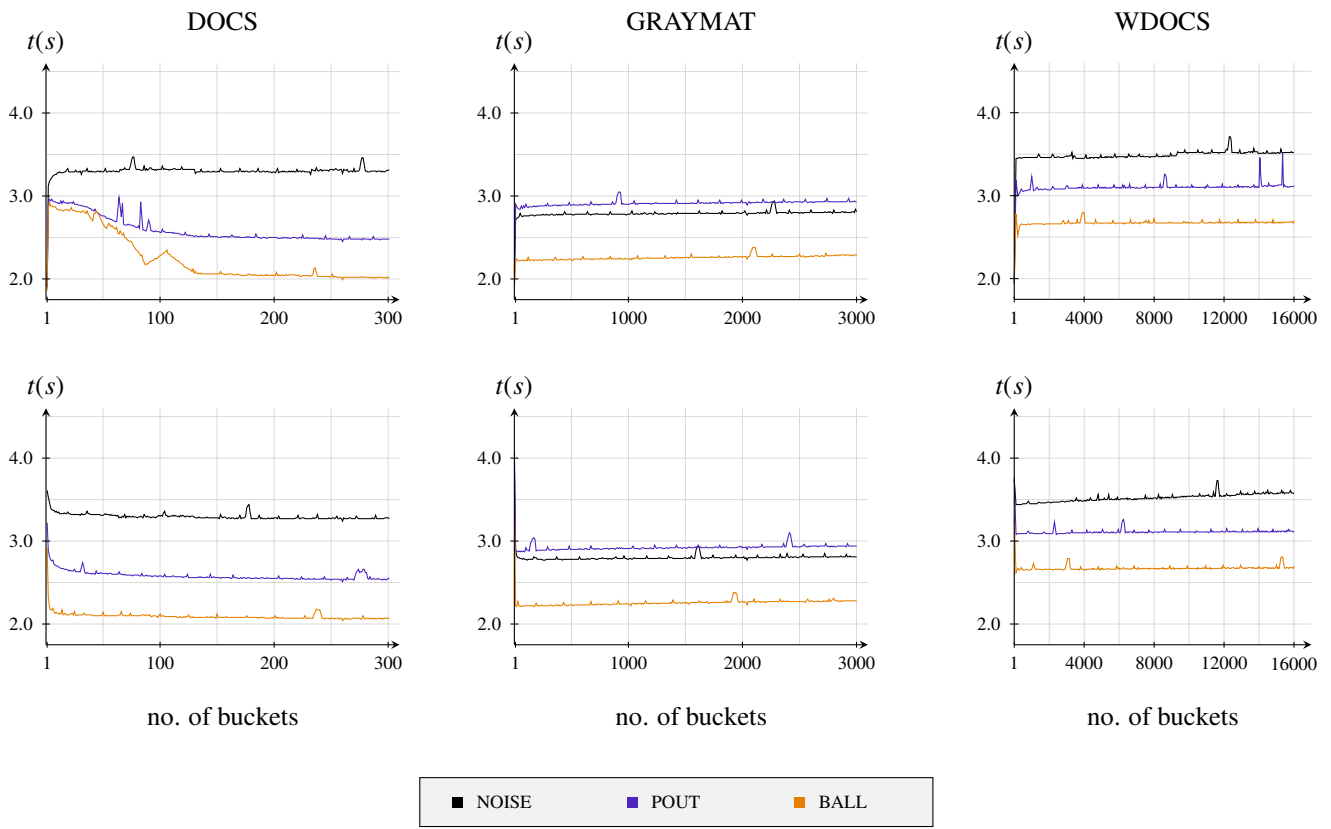

The runtime for different bucket sizes is shown in \figurename~\ref{fig:bucketsizesuntidy}. The DOCS column shows the runtimes for bucket sizes $B \in [1,301]$, the GRAYMAT column for bucket sizes $B \in \{1,11,21,\ldots,3001\}$, and the WDOCS column for bucket sizes $B \in \{1,50,100,\ldots,16000\}$. Each value is the mean runtime over 5 transforms from the same seed.

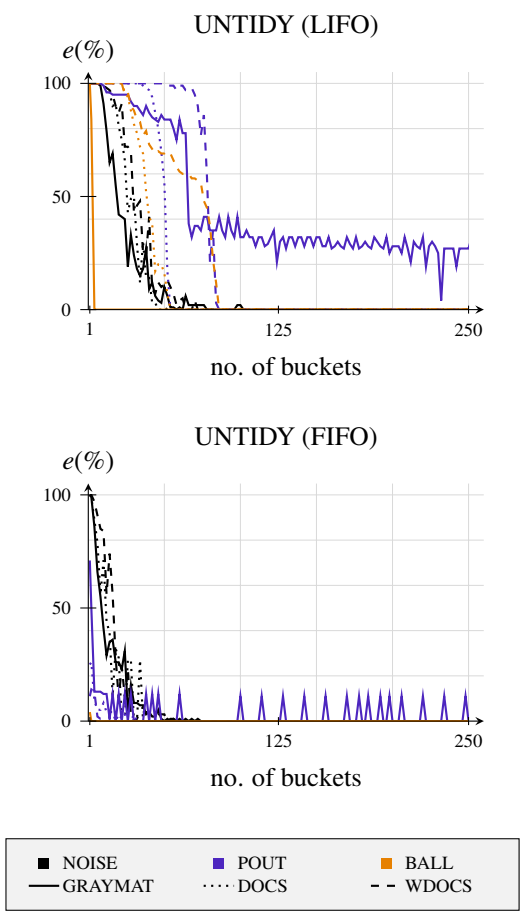
\begin{figure}
  \def\mpscale{1}
  \def\xscale{0.02}
  \def\yscale{3}
  \def\xmin{0}
  \def\xmax{260}
  \def\gridx{260}
  \def\gridy{1.05}
  \def\clipx{250}
  \def\clipy{1}
  \def\ytitle{0.2}
  \def\ymax{1.05}
  \def\ymin{0}

  \begin{minipage}[b]{\mpscale\linewidth}
  \centering
  \begin{tikzpicture}[yscale=\yscale,xscale=\xscale]

    \draw[very thin,color=gray!30,xstep=50,ystep=0.2] (\xmin,\ymin)
    grid +(\gridx,\gridy);
      
    \begin{scope}
      \clip (\xmin,\ymin) rectangle +(\clipx,\clipy);

      \draw[nde] plot file {tab_15_2300_09_110_perc_pix.table};
      \draw[nwe] plot file {tab_15_2300_09_210_perc_pix.table};
      \draw[nge] plot file {tab_15_2300_09_410_perc_pix.table};
      \draw[pde] plot file {tab_15_2300_10_110_perc_pix.table};
      \draw[pwe] plot file {tab_15_2300_10_210_perc_pix.table};
      \draw[pge] plot file {tab_15_2300_10_410_perc_pix.table};
      \draw[bde] plot file {tab_15_2300_11_110_perc_pix.table};
      \draw[bwe] plot file {tab_15_2300_11_210_perc_pix.table};
      \draw[bge] plot file {tab_15_2300_11_410_perc_pix.table};
    \end{scope}

    \draw[->,>=stealth] (\xmin,\ymin) --
    coordinate (x axis mid) (\xmax,\ymin);
    \draw[->,>=stealth] (\xmin,\ymin) -- 
    coordinate (y axis mid) (\xmin,\ymax) node[above] {$e$(\%)};

    \path (x axis mid) +(0,\ymax-\ymin+\ytitle) node {UNTIDY (LIFO)};
    \node[below=0.5cm] at (x axis mid) {no. of buckets};

    \foreach \x in {1, 125, 250}
    \draw[shift={(\x,\ymin)}] (0pt,0.5pt) -- (0pt,-0.5pt)
    node[below] {\tick{$\x$}};

    \foreach \y/\ytext in {0/0, 0.5/50, 1/100}
    \draw[shift={(\xmin,\y)}] (150pt,0pt) -- (-150pt,0pt)
    node[left] {\tick{$\ytext$}};

  \end{tikzpicture}
  \centerline{}
  \end{minipage}
  \hfill
  \begin{minipage}[b]{\mpscale\linewidth}
  \centering
  \begin{tikzpicture}[yscale=\yscale,xscale=\xscale]

    \draw[very thin,color=gray!30,xstep=50,ystep=0.2] (\xmin,\ymin)
    grid +(\gridx,\gridy);
      
    \begin{scope}
      \clip (\xmin,\ymin) rectangle +(\clipx,\clipy);

      \draw[nde] plot file {tab_15_2301_09_110_perc_pix.table};
      \draw[nwe] plot file {tab_15_2301_09_210_perc_pix.table};
      \draw[nge] plot file {tab_15_2301_09_410_perc_pix.table};
      \draw[pde] plot file {tab_15_2301_10_110_perc_pix.table};
      \draw[pwe] plot file {tab_15_2301_10_210_perc_pix.table};
      \draw[pge] plot file {tab_15_2301_10_410_perc_pix.table};
      \draw[bde] plot file {tab_15_2301_11_110_perc_pix.table};
      \draw[bwe] plot file {tab_15_2301_11_210_perc_pix.table};
      \draw[bge] plot file {tab_15_2301_11_410_perc_pix.table};
    \end{scope}

    \draw[->,>=stealth] (\xmin,\ymin) --
    coordinate (x axis mid) (\xmax,\ymin);
    \draw[->,>=stealth] (\xmin,\ymin) -- 
    coordinate (y axis mid) (\xmin,\ymax) node[above] {$e$(\%)};

    \path (x axis mid) +(0,\ymax-\ymin+\ytitle) node {UNTIDY (FIFO)};
    \node[below=0.5cm] at (x axis mid) {no. of buckets};

    \foreach \x in {1, 125, 250}
    \draw[shift={(\x,\ymin)}] (0pt,0.5pt) -- (0pt,-0.5pt)
    node[below] {\tick{$\x$}};

    \foreach \y/\ytext in {0/0, 0.5/50, 1/100}
    \draw[shift={(\xmin,\y)}] (150pt,0pt) -- (-150pt,0pt)
    node[left] {\tick{$\ytext$}};

  \end{tikzpicture}
  \centerline{}
  \end{minipage}
  \begin{minipage}[t]{\linewidth}
  \centering
  \plotlegendextra
  \end{minipage}
\caption{The percent of pixels with erroneous values for different bucket sizes when using the Untidy queue. The top shows the behaviour when using LIFO lists, and the bottom when using FIFO lists.}
\label{fig:untidyerror}
\end{figure}

For GRAYMAT and DOCS we can choose the bucket size required for unique costs (see \tablename~\ref{tab:bucketsize}) and remain confident that the runtime will be kept low. For WDOCS, the required bucket size is more than 18000. \figurename~\ref{fig:bucketsizesuntidy} show that WDOCS has a slight increase in runtime for increasing bucket size. A good choice for bucket size should, thus, be lower than 18000 when considering the runtime. If accuracy of the grey-weighted distance map is not a vital issue, then selecting the number of buckets required for WDOCS can be simplified by quantising the arc weights produced by Equation~(\ref{eq:wdocs}). Here we limit our analysis to non-quantised arc weights. Using fewer buckets than required for unique costs might introduce rounding errors. However, to result in a rounding error two spatially close nodes with different path costs have to end up in the same bucket and get extracted in the wrong order. The probability of this happening is likely to be quite low and depends on both the grey-weighted distance definition and image properties. \figurename~\ref{fig:untidyerror} shows plots of the percent of erroneous values we get in the distance map for different bucket sizes. The plots only show one transform, but it gives an indication that most errors drop off quite fast. To get a more reliable measure on how few buckets can be used without getting an erroneous distance map, we used 5 different seeds from the seed map and recorded the largest bucket size which resulted in an error for each seed. \tablename~\ref{tab:untidyerror} lists the mean and standard deviation over the five seeds for each setting. For DOCS we used bucket sizes $B \in [1,261]$, and for GRAYMAT and WDOCS we used $B \in \{1,3,5,\ldots,2551\}$. The table indicates that bucket sizes much smaller than $18000$ can be chosen for WDOCS without introducing errors in the distance transform. Since \figurename~\ref{fig:bucketsizesuntidy} shows that the choice bucket size for WDOCS is robust with respect to runtime, we use the same number of buckets as was chosen for GRAYMAT (2551) for simplicity.

\begin{table}
\begin{center}
\caption{Largest bucket size resulting in an erroneous grey-weighted distance transform. The value is the mean of five transforms and the bracketed value is the standard deviation.}
\label{tab:untidyerror}
\medskip
\footnotesize{
\begin{tabular}{ll*{3}{r}}
\hline
\noalign{\smallskip}
Image & Alg. & \multicolumn{3}{c}{Bucket size} \\
\cline{3-5}
\noalign{\smallskip}
& & GRAYMAT & DOCS & WDOCS \\
\hline
\noalign{\smallskip}
NOISE & U\sub{L} &  254(84) &   52(0)  &  79(2) \\
      & U\sub{F} &  251(18) &   52(1)  &  77(2) \\
 & & & & \\
POUT  & U\sub{L} & 2455(60) &   52(0)  &  85(0) \\
      & U\sub{F} & 2455(60) &   52(0)  &  78(3) \\
 & & & & \\
BALL  & U\sub{L} & 2149(290) &  52(0)  &  85(0) \\
      & U\sub{F} & 2149(290) &  41(13) &   6(7) \\
\hline
\end{tabular}
}
\end{center}
\end{table}

\comment{
\begin{table}
\begin{center}
\caption{Largest bucket size resulting in an erroneous grey-weighted distance transform.}
\label{tab:untidyerror}
\medskip
\footnotesize{
\begin{tabular}{l*{8}{r}}
\hline
\noalign{\smallskip}
 & \multicolumn{2}{c}{NOISE} & & \multicolumn{2}{c}{POUT} & & \multicolumn{2}{c}{BALL} \\
\cline{2-3}\cline{5-6}\cline{8-9}
\noalign{\smallskip}
 & \multicolumn{1}{c}{LIFO} & \multicolumn{1}{c}{FIFO} & & \multicolumn{1}{c}{LIFO} & \multicolumn{1}{c}{FIFO} & & \multicolumn{1}{c}{LIFO} & \multicolumn{1}{c}{FIFO} \\
\hline
\noalign{\smallskip}
GRAYMAT & 254(84) & 251(18) & & 2455(60) & 2455(60) & &  2149(290) &  2149(290) \\
DOCS    &  52(0) &  52(1) & &   52(0) &   52(0) & & 52(0) & 41(13) \\
WDOCS   &  79(2) &  77(2) & &   85(0) &   78(3) & & 85(0) & 6(7) \\
\hline
\end{tabular}
}
\end{center}
\end{table}
}

\begin{figure*}
  \def\mpscale{0.325}

  \usetikzlibrary[plotmarks]

  \begin{minipage}[t]{\mpscale\linewidth}
  \def\yscale{1.1}
  \def\xscale{0.015}
  \def\xmin{0}
  \def\xmax{310}
  \def\gridx{310}
  \def\ytitle{0.5}
  \def\uymin{1.75}
  \def\uymax{4.60}
  \def\ugridy{2.85}
  \def\myx{1/1,100/100,200/200,300/300}

  \centering
  \begin{tikzpicture}[yscale=\yscale,xscale=\xscale]
    \draw[very thin,color=gray!30,xstep=50,ystep=0.5] (\xmin,\uymin)
    grid +(\gridx,\ugridy);

    \draw[nd] plot file {tab_bs_rt_2400_09_110_rt.table};
    \draw[pd] plot file {tab_bs_rt_2400_10_110_rt.table};
    \draw[bd] plot file {tab_bs_rt_2400_11_110_rt.table};
      
    \draw[->,>=stealth] (\xmin,\uymin) --
    coordinate (x axis mid) (\xmax,\uymin);
    \draw[->,>=stealth] (\xmin,\uymin) --
    coordinate (y axis mid) (\xmin,\uymax) node[above] {$t(s)$};

    \path (x axis mid) +(0,\uymax-\uymin+\ytitle) node {DOCS};
    \node[below=0.7cm] at (x axis mid) {no. of buckets};

    \foreach \x/\xtext in \myx
    \draw[shift={(\x,\uymin)}] (0pt,1pt) -- (0pt,-1pt)
    node[below] {\tick{$\xtext$}};

    \foreach \y in {2.0, 3.0, 4.0}
    \draw[shift={(\xmin,\y)}] (100pt,0pt) -- (-100pt,0pt)
    node[left] {\tick{$\y$}};
  \end{tikzpicture}
  \centerline{}
  \end{minipage}
  \hfill
  \begin{minipage}[t]{\mpscale\linewidth}
  \def\yscale{1.1}
  \def\xscale{0.015}
  \def\xmin{0}
  \def\xmax{310}
  \def\gridx{310}
  \def\ytitle{0.5}
  \def\uymin{1.75}
  \def\uymax{4.60}
  \def\ugridy{2.85}
  \def\myx{1/1,100/1000,200/2000,300/3000}

  \centering
  \begin{tikzpicture}[yscale=\yscale,xscale=\xscale]
    \draw[very thin,color=gray!30,xstep=50,ystep=0.5] (\xmin,\uymin)
    grid +(\gridx,\ugridy);

    \draw[ng] plot file {tab_bs_rt_2400_09_410_rt.table};
    \draw[pg] plot file {tab_bs_rt_2400_10_410_rt.table};
    \draw[bg] plot file {tab_bs_rt_2400_11_410_rt.table};
      
    \draw[->,>=stealth] (\xmin,\uymin) --
    coordinate (x axis mid) (\xmax,\uymin);
    \draw[->,>=stealth] (\xmin,\uymin) --
    coordinate (y axis mid) (\xmin,\uymax) node[above] {$t(s)$};

    \path (x axis mid) +(0,\uymax-\uymin+\ytitle) node {GRAYMAT};
    \node[below=0.7cm] at (x axis mid) {no. of buckets};

    \foreach \x/\xtext in \myx
    \draw[shift={(\x,\uymin)}] (0pt,1pt) -- (0pt,-1pt)
    node[below] {\tick{$\xtext$}};

    \foreach \y in {2.0, 3.0, 4.0}
    \draw[shift={(\xmin,\y)}] (100pt,0pt) -- (-100pt,0pt)
    node[left] {\tick{$\y$}};
  \end{tikzpicture}
  \centerline{}
  \end{minipage}
  \hfill
  \begin{minipage}[t]{\mpscale\linewidth}
  \def\yscale{1.1}
  \def\xscale{0.023}
  \def\xmin{0}
  \def\xmax{170}
  \def\gridx{170}
  \def\ytitle{0.5}
  \def\uymin{1.75}
  \def\uymax{4.60}
  \def\ugridy{2.85}
  \def\myx{0.01/1,40/4000,80/8000,120/12000,160/16000}

  \centering
  \begin{tikzpicture}[yscale=\yscale,xscale=\xscale]
    \draw[very thin,color=gray!30,xstep=20,ystep=0.5] (\xmin,\uymin)
    grid +(\gridx,\ugridy);

    \draw[nw] plot file {tab_bs_rt_2400_09_210_rt.table};
    \draw[pw] plot file {tab_bs_rt_2400_10_210_rt.table};
    \draw[bw] plot file {tab_bs_rt_2400_11_210_rt.table};
      
    \draw[->,>=stealth] (\xmin,\uymin) --
    coordinate (x axis mid) (\xmax,\uymin);
    \draw[->,>=stealth] (\xmin,\uymin) --
    coordinate (y axis mid) (\xmin,\uymax) node[above] {$t(s)$};

    \path (x axis mid) +(0,\uymax-\uymin+\ytitle) node {WDOCS};
    \node[below=0.7cm] at (x axis mid) {no. of buckets};

    \foreach \x/\xtext in \myx
    \draw[shift={(\x,\uymin)}] (0pt,1pt) -- (0pt,-1pt)
    node[below] {\tick{$\xtext$}};

    \foreach \y in {2.0, 3.0, 4.0}
    \draw[shift={(\xmin,\y)}] (70pt,0pt) -- (-70pt,0pt)
    node[left] {\tick{$\y$}};
  \end{tikzpicture}
  \centerline{}
  \end{minipage}
  \begin{minipage}[t]{\linewidth}
  \centering
  \plotlegend
  \end{minipage}
\caption{Runtimes for grey-weighted distance transforms when using different bucket sizes for the hierarchical heap. Note the different scales on the x-axis for the different grey-weighted distance definitions.}
\label{fig:bucketsizeshheap}
\end{figure*}
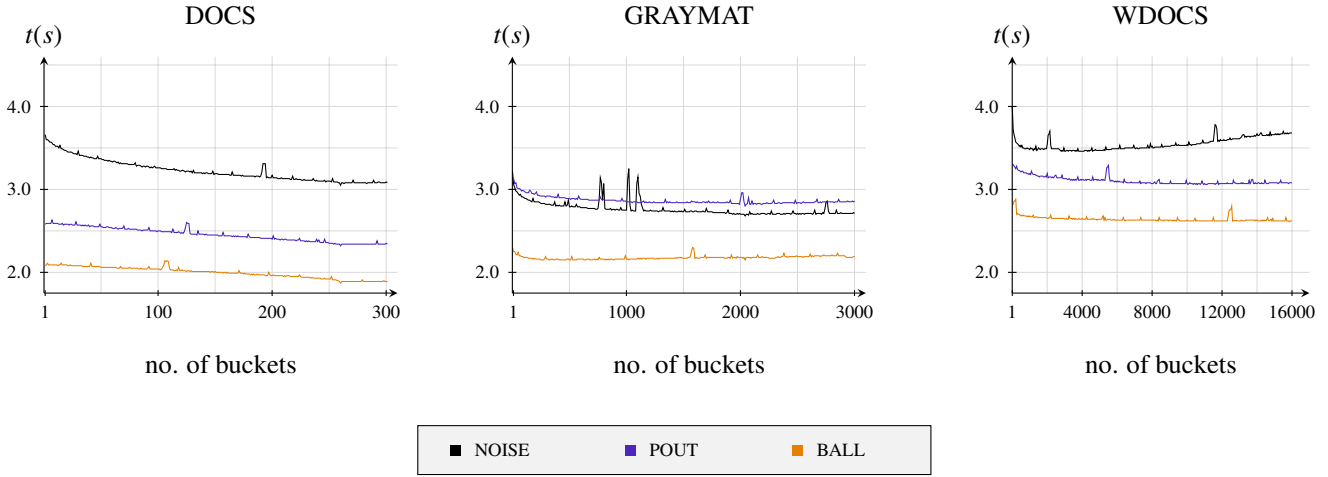

\subsubsection{Hierarchical heap}

Unlike the Untidy queue, the hierarchical heap always returns the node with the lowest path cost from a {\it find-min} operation, independent of bucket size. Therefore, choosing the bucket size for the hierarchical queue can be based on runtime alone. Using less buckets than required for unique path costs will lead to larger heaps since nodes with different costs might have to share buckets. Using more buckets will introduce excessive buckets which cannot get any nodes assigned to them. This will only make the circular array longer and take more time to sift through. The runtime for different bucket sizes is shown in \figurename~\ref{fig:bucketsizeshheap}.

The DOCS definition shows a decline in runtime which levels out around 250, which makes 261, the bucket size required for unique path costs, the most reasonable choice of bucket size for the hierarchical heap when using DOCS. The GRAYMAT definition behaves similarly to the DOCS definition for NOISE and POUT, decreasing in runtime and levelling out when getting close to the bucket size required for unique path costs. For BALL the runtime seems to increase slightly with bucket size. However, the difference in runtime for small versus large bucket sizes is so low that the behaviour for NOISE and POUT is more important. Therefore, the bucket size required for unique path costs (2551) is a good choice for the hierarchical heap when using GRAYMAT. Since \figurename~\ref{fig:bucketsizeshheap} indicates that the choice of bucket size for WDOCS is robust with respect to runtime, except for NOISE where it increases slightly for high bucket sizes, we use the same number of buckets as was chosen for GRAYMAT (2551) for simplicity.


\subsubsection{Cost spread}\label{sec:costspread}

\begin{figure*}
  \pgfmathsetlength{\imagewidth}{7cm} 

  \newcommand{\imtexth}[1]{\textbf{\textsf{\textcolor{white}{\scriptsize{#1}}}}}
  \newcommand{\imtext}[1]{\textbf{\textsf{\textcolor{white}{\scriptsize{#1}}}}}

  \def\mpscale{0.45}
  \def\imagescale{0.0116666667}
  \def\xval{575}
  \def\xtext{520}
  \def\ymax{24}
  \def\ymin{46}
  \def\ymean{68}
  \def\ystd{90}

  \def\smax{max}
  \def\smin{min}
  \def\smean{mean}
  \def\sstd{std}

  \def\lw{0.12cm}
  \def\xscale{0.45}
  \def\yscale{0.05}

  \newcommand{\hist}{
    \begin{scope}
      \clip (-1.5,-10) rectangle (13,35);
      
      \foreach \y in {0, 10, 20, 30}
      \draw[very thin,color=gray!30] (0,\y) -- (13,\y);

      \foreach \y in {0, 10, 20, 30}
      \draw (0.1,\y) -- (-0.1,\y) node[left] {\scriptsize{\y}};

      \foreach \x/\xtext in {0.75, 1.75, 2.75, 3.75, 4.75, 5.75, 6.75,
	                     7.75, 8.75, 9.75, 10.75, 11.75}
      \draw (\x,0) -- (\x,-3);
    \end{scope}
      
    \draw[->,>=stealth] (0,0) -- coordinate (x axis mid) (13.5,0);
    \draw[->,>=stealth] (0,0) -- coordinate (y axis mid) (0,35);
  }

  \newcommand{\legend}[1]{%
    \node[anchor=north west, rectangle, draw=black, fill=gray!10,
      inner sep=5pt, rounded corners] at (8,35) {\footnotesize{#1}};
  }

  \newcommand{\legendhead}{%
    \def\legendborderleft{0.5}
    \def\legendborderright{2.5}
    \def\legendbordertop{2}
    \def\legendborderbottom{2}


    \node (leg1) at (3,45) {};
    \path (leg1) +(4,0) node (leg2) {};
    \path (leg2) +(2.7,0) node (leg3) {};
  
    \path (leg1.north west)+(-\legendborderleft,\legendbordertop) node (a) {};
    \path (leg3.south east)+(\legendborderright,-\legendborderbottom) node (b) {};

    \path[draw=black,fill=gray!10] (a) rectangle (b);    

    \node[anchor=west,font=\scriptsize] at (leg1) {GRAYMAT};
    \node[anchor=west,font=\scriptsize] at (leg2) {DOCS};
    \node[anchor=west,font=\scriptsize] at (leg3) {WDOCS};

    \node[rectangle,draw=cg,fill=cg,inner sep=2pt] at (leg1.west) {};
    \node[rectangle,draw=cd,fill=cd,inner sep=2pt] at (leg2.west) {};
    \node[rectangle,draw=cw,fill=cw,inner sep=2pt] at (leg3.west) {};
  
  }

  \begin{minipage}[b]{\mpscale\linewidth}

  \begin{minipage}[b]{\linewidth}
  \centering
  \centerline{NOISE}
  \centerline{}
  \begin{tikzpicture}[xscale=\imagescale,yscale=-\imagescale]
    \node[anchor=north west,inner sep=0pt,outer sep=0pt] at (-0.5,-0.5)
       {\includegraphics[width=\imagewidth]{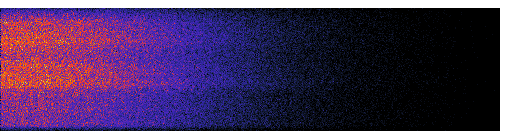}};
       
    \node[anchor=north east] at (\xval,0) {\imtexth{GRAYMAT}};
    \node[anchor=north east] at (\xval,\ymax) {\imtext{74\%}};
    \node[anchor=north east] at (\xval,\ymin) {\imtext{42\%}};
    \node[anchor=north east] at (\xval,\ymean) {\imtext{69\%}};
    \node[anchor=north east] at (\xval,\ystd) {\imtext{4\%}};

    \node[anchor=north east] at (\xtext,\ymax) {\imtext{\smax}};
    \node[anchor=north east] at (\xtext,\ymin) {\imtext{\smin}};
    \node[anchor=north east] at (\xtext,\ymean) {\imtext{\smean}};
    \node[anchor=north east] at (\xtext,\ystd) {\imtext{\sstd}};

  \end{tikzpicture}
  \centerline{}
  \end{minipage}
  \begin{minipage}[b]{\linewidth}
  \centering
  \begin{tikzpicture}[xscale=\imagescale,yscale=-\imagescale]
    \node[anchor=north west,inner sep=0pt,outer sep=0pt] at (-0.5,-0.5)
       {\includegraphics[width=\imagewidth]{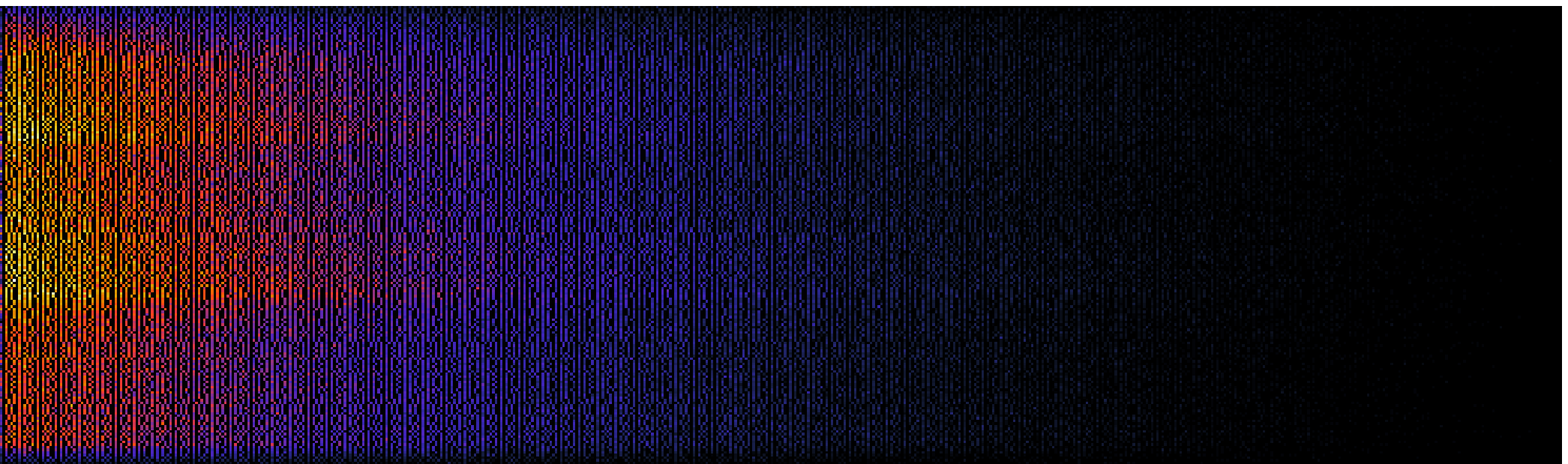}};

    \node[anchor=north east] at (\xval,0) {\imtexth{DOCS}};
    \node[anchor=north east] at (\xval,\ymax) {\imtext{39\%}};
    \node[anchor=north east] at (\xval,\ymin) {\imtext{25\%}};
    \node[anchor=north east] at (\xval,\ymean) {\imtext{36\%}};
    \node[anchor=north east] at (\xval,\ystd) {\imtext{2\%}};

    \node[anchor=north east] at (\xtext,\ymax) {\imtext{\smax}};
    \node[anchor=north east] at (\xtext,\ymin) {\imtext{\smin}};
    \node[anchor=north east] at (\xtext,\ymean) {\imtext{\smean}};
    \node[anchor=north east] at (\xtext,\ystd) {\imtext{\sstd}};

  \end{tikzpicture}
  \centerline{}
  \end{minipage}
  \begin{minipage}[b]{\linewidth}
  \centering
  \begin{tikzpicture}[xscale=\imagescale,yscale=-\imagescale]
    \node[anchor=north west,inner sep=0pt,outer sep=0pt] at (-0.5,-0.5)
       {\includegraphics[width=\imagewidth]{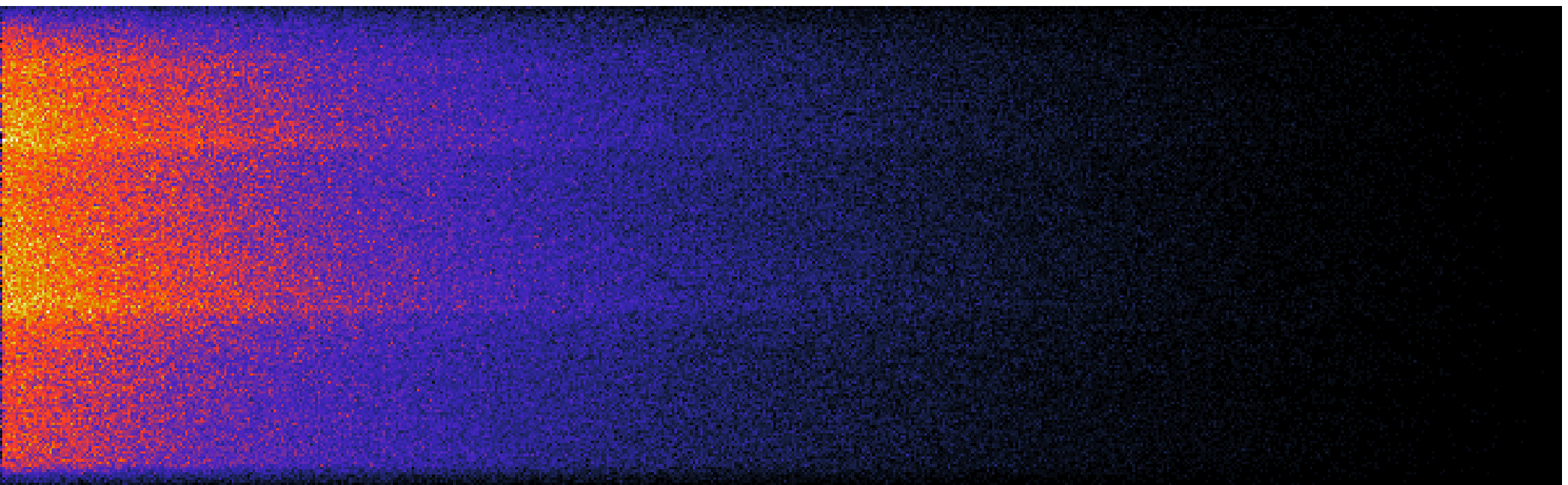}};

    \node[anchor=north east] at (\xval,0) {\imtexth{WDOCS}};
    \node[anchor=north east] at (\xval,\ymax) {\imtext{83\%}};
    \node[anchor=north east] at (\xval,\ymin) {\imtext{43\%}};
    \node[anchor=north east] at (\xval,\ymean) {\imtext{77\%}};
    \node[anchor=north east] at (\xval,\ystd) {\imtext{4\%}};

    \node[anchor=north east] at (\xtext,\ymax) {\imtext{\smax}};
    \node[anchor=north east] at (\xtext,\ymin) {\imtext{\smin}};
    \node[anchor=north east] at (\xtext,\ymean) {\imtext{\smean}};
    \node[anchor=north east] at (\xtext,\ystd) {\imtext{\sstd}};
  \end{tikzpicture}
  \centerline{}
  \end{minipage}
  \end{minipage}
  \hfill
  \begin{minipage}[b]{\mpscale\linewidth}
  \begin{minipage}[b]{\linewidth}
  \centering
  \centerline{POUT}
  \centerline{}
  \begin{tikzpicture}[xscale=\imagescale,yscale=-\imagescale]
    \node[anchor=north west,inner sep=0pt,outer sep=0pt] at (-0.5,-0.5)
       {\includegraphics[width=\imagewidth]{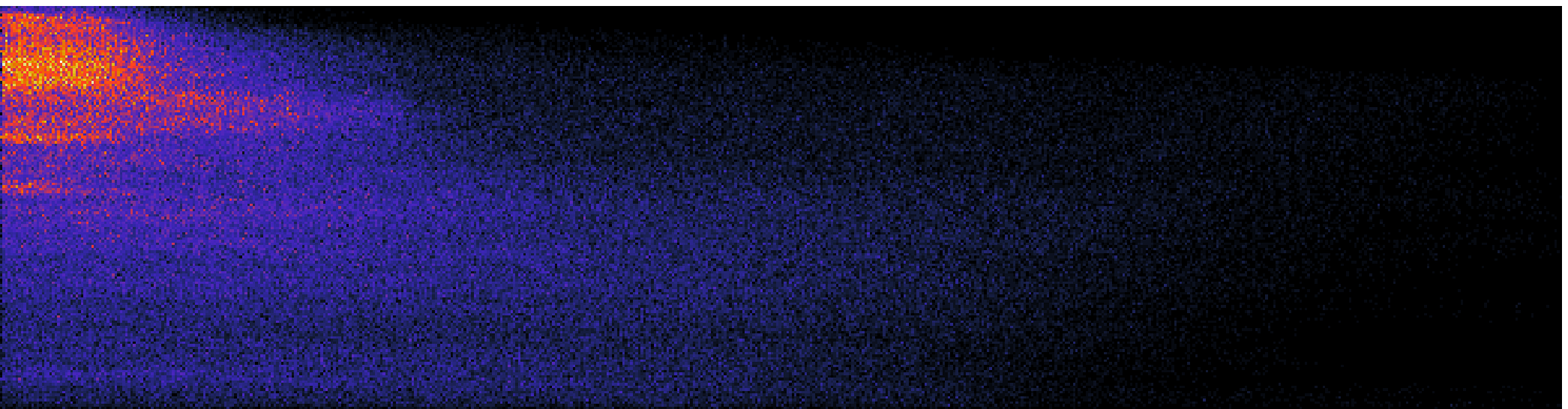}};

    \node[anchor=north east] at (\xval,0) {\imtexth{GRAYMAT}};
    \node[anchor=north east] at (\xval,\ymax) {\imtext{84\%}};
    \node[anchor=north east] at (\xval,\ymin) {\imtext{12\%}};
    \node[anchor=north east] at (\xval,\ymean) {\imtext{69\%}};
    \node[anchor=north east] at (\xval,\ystd) {\imtext{17\%}};

    \node[anchor=north east] at (\xtext,\ymax) {\imtext{\smax}};
    \node[anchor=north east] at (\xtext,\ymin) {\imtext{\smin}};
    \node[anchor=north east] at (\xtext,\ymean) {\imtext{\smean}};
    \node[anchor=north east] at (\xtext,\ystd) {\imtext{\sstd}};
  \end{tikzpicture}
  \centerline{}
  \end{minipage}
  \begin{minipage}[b]{\linewidth}
  \centering
  \begin{tikzpicture}[xscale=\imagescale,yscale=-\imagescale]
    \node[anchor=north west,inner sep=0pt,outer sep=0pt] at (-0.5,-0.5)
       {\includegraphics[width=\imagewidth]{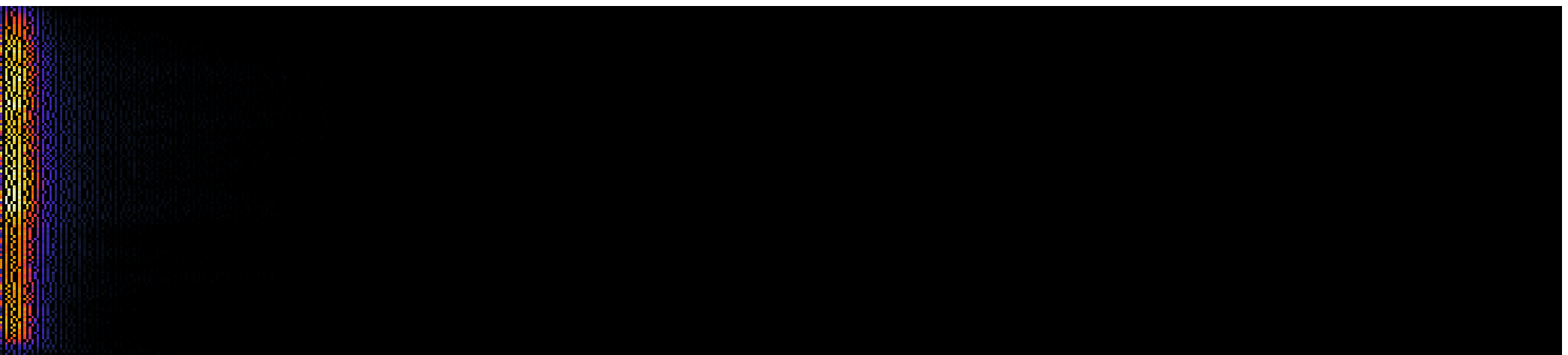}};

    \node[anchor=north east] at (\xval,0) {\imtexth{DOCS}};
    \node[anchor=north east] at (\xval,\ymax) {\imtext{9\%}};
    \node[anchor=north east] at (\xval,\ymin) {\imtext{2\%}};
    \node[anchor=north east] at (\xval,\ymean) {\imtext{5\%}};
    \node[anchor=north east] at (\xval,\ystd) {\imtext{2\%}};

    \node[anchor=north east] at (\xtext,\ymax) {\imtext{\smax}};
    \node[anchor=north east] at (\xtext,\ymin) {\imtext{\smin}};
    \node[anchor=north east] at (\xtext,\ymean) {\imtext{\smean}};
    \node[anchor=north east] at (\xtext,\ystd) {\imtext{\sstd}};
  \end{tikzpicture}
  \centerline{}
  \end{minipage}
  \begin{minipage}[b]{\linewidth}
  \centering
  \begin{tikzpicture}[xscale=\imagescale,yscale=-\imagescale]
    \node[anchor=north west,inner sep=0pt,outer sep=0pt] at (-0.5,-0.5)
       {\includegraphics[width=\imagewidth]{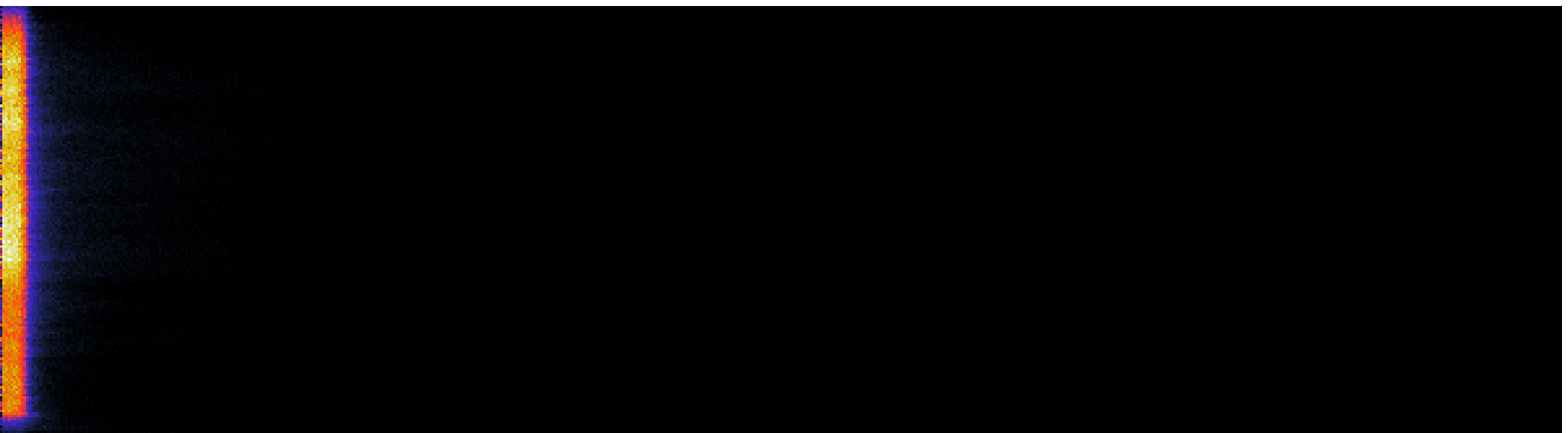}};

    \node[anchor=north east] at (\xval,0) {\imtexth{WDOCS}};
    \node[anchor=north east] at (\xval,\ymax) {\imtext{16\%}};
    \node[anchor=north east] at (\xval,\ymin) {\imtext{3\%}};
    \node[anchor=north east] at (\xval,\ymean) {\imtext{10\%}};
    \node[anchor=north east] at (\xval,\ystd) {\imtext{4\%}};

    \node[anchor=north east] at (\xtext,\ymax) {\imtext{\smax}};
    \node[anchor=north east] at (\xtext,\ymin) {\imtext{\smin}};
    \node[anchor=north east] at (\xtext,\ymean) {\imtext{\smean}};
    \node[anchor=north east] at (\xtext,\ystd) {\imtext{\sstd}};
  \end{tikzpicture}
  \centerline{}
  \centerline{}
  \centerline{}
  \end{minipage}
  \end{minipage}
  \begin{minipage}[b]{\mpscale\linewidth}
  \begin{minipage}[b]{\linewidth}
  \centering
  \centerline{}
  \centerline{}
  \centerline{BALL}
  \centerline{}
  \begin{tikzpicture}[xscale=\imagescale,yscale=-\imagescale]
    \node[anchor=north west,inner sep=0pt,outer sep=0pt] at (-0.5,-0.5)
       {\includegraphics[width=\imagewidth]{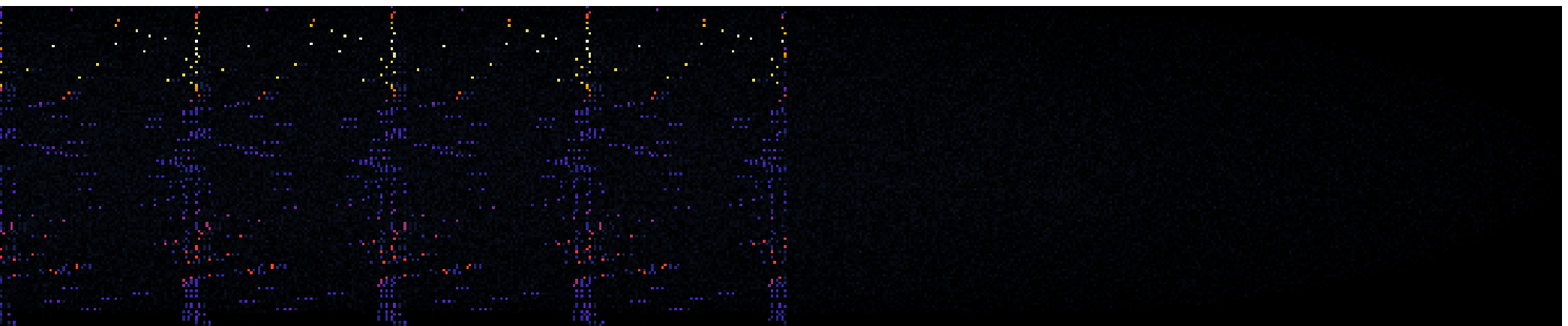}};

    \node[anchor=north east] at (\xval,0) {\imtexth{GRAYMAT}};
    \node[anchor=north east] at (\xval,\ymax) {\imtext{85\%}};
    \node[anchor=north east] at (\xval,\ymin) {\imtext{2\%}};
    \node[anchor=north east] at (\xval,\ymean) {\imtext{67\%}};
    \node[anchor=north east] at (\xval,\ystd) {\imtext{18\%}};

    \node[anchor=north east] at (\xtext,\ymax) {\imtext{\smax}};
    \node[anchor=north east] at (\xtext,\ymin) {\imtext{\smin}};
    \node[anchor=north east] at (\xtext,\ymean) {\imtext{\smean}};
    \node[anchor=north east] at (\xtext,\ystd) {\imtext{\sstd}};
  \end{tikzpicture}
  \centerline{}
  \end{minipage}
  \begin{minipage}[b]{\linewidth}
  \centering
  \begin{tikzpicture}[xscale=\imagescale,yscale=-\imagescale]
    \node[anchor=north west,inner sep=0pt,outer sep=0pt] at (-0.5,-0.5)
       {\includegraphics[width=\imagewidth]{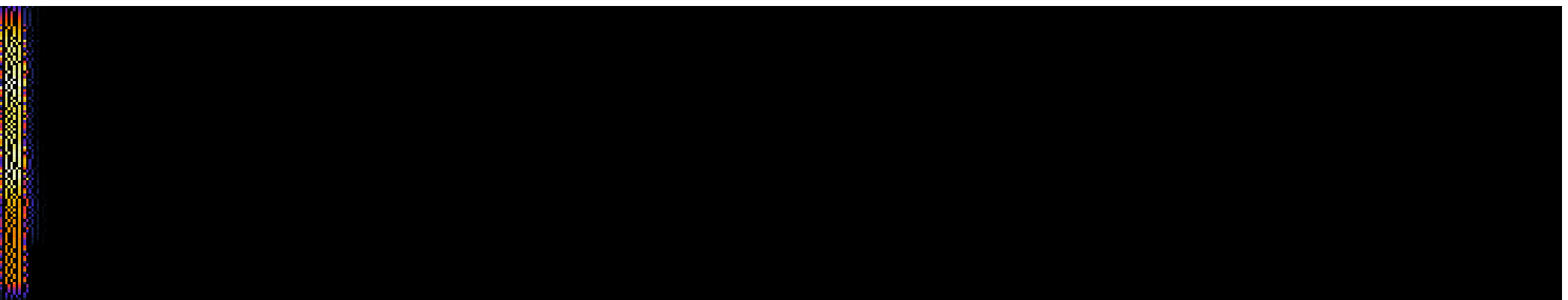}};

    \node[anchor=north east] at (\xval,0) {\imtexth{DOCS}};
    \node[anchor=north east] at (\xval,\ymax) {\imtext{1\%}};
    \node[anchor=north east] at (\xval,\ymin) {\imtext{1\%}};
    \node[anchor=north east] at (\xval,\ymean) {\imtext{1\%}};
    \node[anchor=north east] at (\xval,\ystd) {\imtext{0\%}};

    \node[anchor=north east] at (\xtext,\ymax) {\imtext{\smax}};
    \node[anchor=north east] at (\xtext,\ymin) {\imtext{\smin}};
    \node[anchor=north east] at (\xtext,\ymean) {\imtext{\smean}};
    \node[anchor=north east] at (\xtext,\ystd) {\imtext{\sstd}};
  \end{tikzpicture}
  \centerline{}
  \end{minipage}
  \begin{minipage}[b]{\linewidth}
  \centering
  \begin{tikzpicture}[xscale=\imagescale,yscale=-\imagescale]
    \node[anchor=north west,inner sep=0pt,outer sep=0pt] at (-0.5,-0.5)
       {\includegraphics[width=\imagewidth]{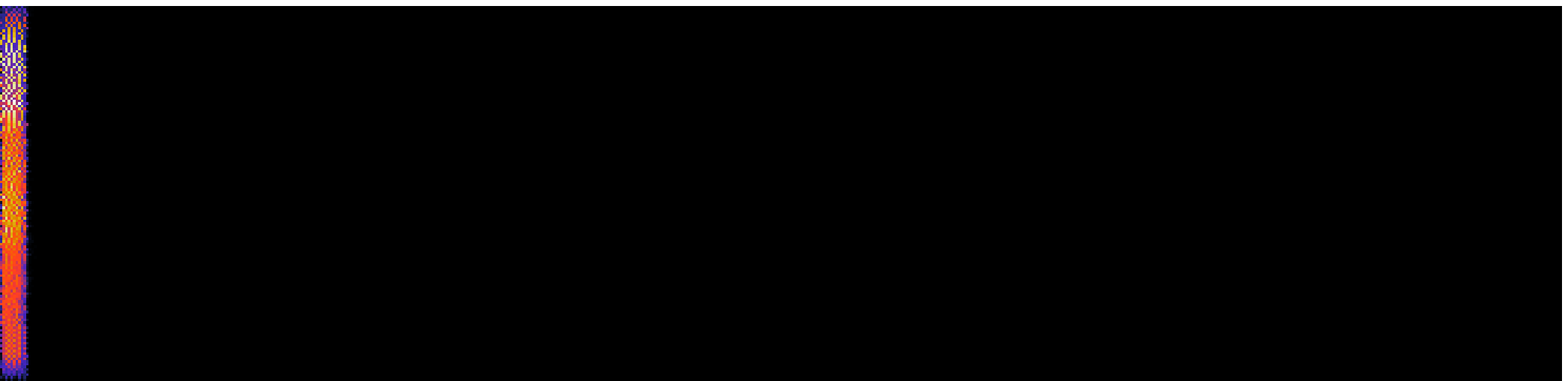}};

    \node[anchor=north east] at (\xval,0) {\imtexth{WDOCS}};
    \node[anchor=north east] at (\xval,\ymax) {\imtext{2\%}};
    \node[anchor=north east] at (\xval,\ymin) {\imtext{1\%}};
    \node[anchor=north east] at (\xval,\ymean) {\imtext{2\%}};
    \node[anchor=north east] at (\xval,\ystd) {\imtext{0\%}};

    \node[anchor=north east] at (\xtext,\ymax) {\imtext{\smax}};
    \node[anchor=north east] at (\xtext,\ymin) {\imtext{\smin}};
    \node[anchor=north east] at (\xtext,\ymean) {\imtext{\smean}};
    \node[anchor=north east] at (\xtext,\ystd) {\imtext{\sstd}};
  \end{tikzpicture}
  \centerline{}
  \centerline{}
  \centerline{}
  \centerline{}
  \end{minipage}
  \end{minipage}
  \hfill
  \begin{minipage}[b]{\mpscale\linewidth}
  \begin{minipage}[b]{\linewidth}
  \begin{tikzpicture}[xscale=\xscale,yscale=\yscale]
    \hist
    \begin{scope}
      \clip (-1.5,-10) rectangle (13,35);
 
      \draw[ycomb,color=cg,line width=\lw]
        plot coordinates {(1,16) (2,15) (3,13) (4,10) (5,7) (6,5) (7,3) (8,1)
	                  (9,1) (10,0) (11,0) (12,0)};
      \draw[ycomb,color=cd,line width=\lw]
        plot coordinates {(1.25,30) (2.25,23) (3.25,17) (4.25,13) (5.25,9)
	                  (6.25,6) (7.25,4) (8.25,2) (9.25,1) (10.25,1)
	                  (11.25,0) (12.25,0)};
      \draw[ycomb,color=cw,line width=\lw]
        plot coordinates {(1.5,34) (2.5,25) (3.5,19) (4.5,14) (5.5,10) (6.5,7)
	                  (7.5,5) (8.5,3) (9.5,2) (10.5,1) (11.5,0) (12.5,0)};
    \end{scope}

    \draw[->,>=stealth] (0,0) -- coordinate (x axis mid) (13.5,0);

    \legendhead
    \legend{NOISE}
  \end{tikzpicture}
  \end{minipage}
  \begin{minipage}[b]{\linewidth}
  \begin{tikzpicture}[xscale=\xscale,yscale=\yscale]
    \hist
    \clip (-1.5,-10) rectangle (13,35);
 
    \draw[ycomb,color=cg,line width=\lw]
      plot coordinates {(1,16) (2,12) (3,8) (4,6) (5,4) (6,4) (7,3) (8,2) (9,1)
	                (10,1) (11,0) (12,0)};
    \draw[ycomb,color=cd,line width=\lw]
      plot coordinates {(1.25,46) (2.25,1) (3.25,0) (4.25,0) (5.25,0) (6.25,0)
	                (7.25,0) (8.25,0) (9.25,0) (10.25,0) (11.25,0)
	                (12.25,0)};
    \draw[ycomb,color=cw,line width=\lw]
      plot coordinates {(1.5,49) (2.5,1) (3.5,0) (4.5,0) (5.5,0) (6.5,0)
	                (7.5,0) (8.5,0) (9.5,0) (10.5,0) (11.5,0) (12.5,0)};

    \draw[->,>=stealth] (0,0) -- coordinate (x axis mid) (13.5,0);

    \legend{POUT}
  \end{tikzpicture}
  \end{minipage}
  \begin{minipage}[b]{\linewidth}
  \begin{tikzpicture}[xscale=\xscale,yscale=\yscale]
    \hist
    \clip (-1.5,-23) rectangle (13,35);
 
    \draw[ycomb,color=cg,line width=\lw]
      plot coordinates {(1,4) (2,5) (3,4) (4,4) (5,5) (6,4) (7,2) (8,1) (9,1)
	                (10,1) (11,1) (12,0)};
    \draw[ycomb,color=cd,line width=\lw]
      plot coordinates {(1.25,31) (2.25,0) (3.25,0) (4.25,0) (5.25,0) (6.25,0)
	                (7.25,0) (8.25,0) (9.25,0) (10.25,0) (11.25,0)
	                (12.25,0)};
    \draw[ycomb,color=cw,line width=\lw]
      plot coordinates {(1.5,38) (2.5,0) (3.5,0) (4.5,0) (5.5,0) (6.5,0)
	                (7.5,0) (8.5,0) (9.5,0) (10.5,0) (11.5,0) (12.5,0)};

    \foreach \x/\xtext in {0.75/1-50, 1.75/51-100, 2.75/101-150, 3.75/151-200,
                           4.75/201-250, 5.75/251-300, 6.75/301-350,
			   7.75/351-400, 8.75/401-450, 9.75/451-500,
			   10.75/501-550, 11.75/551-600 }
      \draw (\x,0) -- (\x,-3)
      node[below,rotate=90,shift={(-0.4,0)}] {\scriptsize{\xtext}};

    \draw[->,>=stealth] (0,0) -- coordinate (x axis mid) (13.5,0);

    \legend{BALL}
  \end{tikzpicture}
  \end{minipage}
  \end{minipage}
\caption{Bucket queue spread for each grey-weighted distance definition during a transform when using 600 buckets. Each pixel row shows a snapshot of how many nodes occupy each bucket at every 10000th iteration (from top to bottom). The first pixel in every row is the bucket with the lowest cost nodes, and the colour ranges from dark blue (empty bucket) to red (bucket with the highest number of nodes). The bar diagrams show the mean spread in number of nodes per bucket for each image and grey-weighted distance definition.}
\label{fig:nodemap}
\end{figure*}

Another interesting aspect of the bucket queues is the spread of the queue over time. The spread shows how many nodes occupy each bucket at a specific time during the calculation of the grey-weighted distance transform. \figurename~\ref{fig:nodemap} shows the spread for a transform, using bucket size $B = 600$, for each image type and each grey-weighted distance definition. The queue was sampled at every 10'000th iteration, and each row represents the state of the queue at the time (from top to bottom), with every pixel being one bucket. The values ranges from dark blue (empty bucket) to red (maximum number of nodes). The \texttt{max}, \texttt{min}, \texttt{mean}, and \texttt{std} percentages are statistics on the number of non-empty buckets. The figure also shows the mean spread as a bar diagram for each image.

It is clear that the noisier image result in a wider range of arc weights, giving rise to a wider spread, while both POUT and BALL give a low spread for DOCS and WDOCS. This means that for DOCS and WDOCS, a large increase in bucket size will only have a small effect on the number of buckets used, which can be one of the explanations why the runtime does not vary much between different bucket sizes.

The thin spread for DOCS and WDOCS can be utilised by removing the unused part of the circular queue to get an increased spread. A wider spread will likely lead to shallower heaps for small bucket sizes in the hierarchical heap, and decreased rounding errors for small bucket sizes in the Untidy queue, which in turn can lead to shorter runtimes. Note however that a more thorough statistical analysis of the spread needs to be done to draw any conclusions on how much the circular array can be shortened for the different cases. Such an analysis is out of scope for this article.

\subsection{Tests on 2D images}\label{sec:2dtests}

Each of the 49 points in the test point grid in \figurename~\ref{fig:testimages2d}~(c) was used as a seed in a seeded grey-weighted distance transform of each of the 2D images. \tablename~\ref{tab:runtimes2d-960} shows the average runtimes in seconds. When running the experiments, it became apparent that the recursive propagation algorithm (labels P\sub{L} and P\sub{LA}) expands pixels in an order which is extremely ineffective for grey-weighted distance computations. The pixel removed in Step 4 in Algorithm~\ref{alg:propagation} is always the pixel most recently added to the list $L$. This has the effect that the algorithm starts by expanding pixels along a path from the seed throughout the image, calculating numerous grey-weighted values based on the one seed neighbour at the beginning of the path. Since grey-weighted distances is monotonically increasing from the seed and depend on the local neighbourhood, expanding its neighbours (which is done at some point when the recursive propagation passes by on its way back for P\sub{L} or has expanded all subsequent pixels in the image for P\sub{LA}) will most likely make the initial path calculation redundant. For example, the recursive propagation algorithm P\sub{L} took 4031 seconds to compute one GRAYMAT on a 240$\times$291 version of \mbox{NOISE}, visiting each pixel more than 37000 times on average. The chamfer algorithm, in comparison, took 8.16 seconds on average (standard deviation $\sigma = 1.59$) and visited each pixel an average of 42.6 times. Because of this, the recursive propagation algorithms were excluded from any further experiments and do not show up in the table. 

In~\tablename~\ref{tab:runtimes2d-960} we see that the overall best performers were the label-setting algorithm using the binary heap with pointer array (H\sub{A}), the Dial queue (D\sub{L} and D\sub{F}), the Untidy queue (U\sub{L} and U\sub{F}), and the hierarchical heap with pointer array (HH\sub{A}). It is apparent that the label-correcting algorithms are a poor choice due to their habit of visiting each node multiple times. The binary heap with pointer array was the fastest in two out of the nine cases; when calculating the WDOCS transform on the NOISE and POUT images. The hierarchical heap without pointer array was the fastest in two cases; when calculating the GRAYMAT and DOCS transform on the BALL image. The hierarchical heap with pointer array was the fastest in the rest of the cases and tied with the binary heap with pointer array when calculating the WDOCS transform in the NOISE image. The hash functions clearly have too much overhead and the algorithms with hash tables perform worse than they do without any helper structures.

It is apparent that the Fibonacci heap performs worse than the binary heap. The lower, amortised, time complexity of the Fibonacci heap does not beat the binary heap for grey-weighted distance computations. This indicates that the number of {\it extract-min} and {\it delete} operations is high relative to the number of other operations performed on the heap.

The differences in runtime between the Dial queue and the Untidy queue are small, which was anticipated considering their similar structure. The performance of the Dial queue is slightly better than the Untidy queue, most likely due to the extra computation required by the Untidy queue to determine which bucket to put a node (Equation~(\ref{eq:bucket})). We can conclude that a LIFO list in the Dial or Untidy queue results in a slightly better runtime than a FIFO list. This is likely to stem from the fact that requires a LIFO list requires fewer operations than a FIFO list.



\begin{table*}
\begin{center}
\caption{Running time (in seconds) for grey-weighted distance transforms on the NOISE, POUT, and BALL images. The best times are marked with black colour, times within 10\% of the best with dark grey, and times within 40\% of best with light grey.}
\label{tab:runtimes2d-960}
\medskip
\footnotesize{
\begin{tabular}{l*{11}{r}}
\hline
\noalign{\smallskip}
 & \multicolumn{3}{c}{NOISE} & & \multicolumn{3}{c}{POUT} & & \multicolumn{3}{c}{BALL} \\
\cline{2-4}\cline{6-8}\cline{10-12}
\noalign{\smallskip}
 & GRAYMAT & DOCS & WDOCS & & GRAYMAT & DOCS & WDOCS & & GRAYMAT & DOCS & WDOCS \\
\hline
\noalign{\smallskip}
C & 697.01  & 683.31 & 703.59 &  & 223.24 & 166.77 & 289.24 &  & 22.02 & 60.42 & 246.48 \\ 
P\sub{F}    & 23.09 & 19.81 & 21.86 &  & 156.62 & 14.73 & 13.98 &  & 27.16 & 12.33 & 12.90 \\ 
P\sub{FA}   & 107.14 & 80.43 & 93.48 &  & 229.80 & 16.62 & 10.95 &  & 46.08 & 4.01 & 4.14 \\ 
H           & \forty{3.05} & 3.54 & 3.89 &  & \forty{3.19} & \forty{2.58} & \forty{3.32} &  & \ten{2.36} & \forty{2.09} & \forty{2.74} \\ 
H\sub{A}    & \ten{2.54} & \ten{2.54} & \best{2.65} &  & \ten{2.50} & \ten{2.33} & \best{2.50} &  & \ten{2.33} & \forty{2.24} & \ten{2.39} \\ 
H\sub{LIN}  & 4.02 & 4.23 & 4.35 &  & 3.93 & 3.62 & 3.67 &  & 3.41 & 3.37 & 3.41 \\ 
H\sub{SUM}  & 3.92 & 4.14 & 4.26 &  & 3.71 & 3.49 & \forty{3.50} &  & 3.25 & 3.13 & \forty{3.22} \\ 
H\sub{PROD} & 4.42 & 4.69 & 4.82 &  & 4.26 & 3.91 & 3.93 &  & 3.74 & 3.64 & 3.72 \\ 
H\sub{XOR}  & 4.55 & 4.88 & 5.10 &  & 4.31 & 3.95 & 3.93 &  & 3.61 & 3.59 & 3.61 \\ 
F           & 4.47 & 5.08 & 5.72 &  & 4.57 & 3.53 & 4.79 &  & 3.33 & 2.82 & 3.91 \\ 
F\sub{A}    & 3.55 & 3.46 & 3.74 &  & 3.42 & 3.09 & \forty{3.45} &  & 3.18 & 2.94 & \forty{3.27} \\ 
F\sub{SUM}  & 5.12 & 5.02 & 5.32 &  & 4.66 & 4.35 & 4.59 &  & 4.22 & 3.93 & 4.15 \\ 
D\sub{L}    & - & \forty{3.22} & - &  & - & \forty{2.43} & - &  & - & \ten{1.98} & - \\ 
D\sub{F}    & - & \forty{3.29} & - &  & - & \forty{2.49} & - &  & - & \ten{2.02} & - \\ 
D\sub{LA}   & - & \forty{2.65} & - &  & - & \forty{2.58} & - &  & - & \forty{2.13} & - \\ 
U\sub{L}    & \forty{2.82} & \forty{3.28} & \forty{3.51} &  & \forty{2.96} & \forty{2.48} & \forty{3.07} &  & \ten{2.28} & \ten{2.02} & \ten{2.60} \\ 
U\sub{F}    & \forty{2.83} & \forty{3.29} & \forty{3.52} &  & \forty{2.96} & \forty{2.54} & \forty{3.09} &  & \ten{2.27} & \ten{2.05} & \ten{2.60} \\ 
U\sub{LA}   & \ten{2.47} & \forty{2.74} & \ten{2.68} &  & \ten{2.49} & \forty{2.63} & \ten{2.62} &  & \ten{2.27} & \forty{2.17} & \ten{2.47} \\ 
HH          & \forty{2.75} & \forty{3.14} & \forty{3.57} &  & \forty{2.88} & \ten{2.35} & \forty{3.20} &  & \best{2.20} & \best{1.89} & \forty{2.63} \\ 
HH\sub{A}   & \best{2.39} & \best{2.38} & \best{2.65} &  & \best{2.39} & \best{2.18} & \ten{2.51} &  & \ten{2.21} & \ten{2.05} & \best{2.36} \\ 
\hline
\end{tabular}
}
\end{center}
\end{table*}

\subsection{Tests on 3D images}\label{sec:3dtests}

\subsubsection{CT}\label{sec:3dtests_ct}

Here we used the CT image with size 256$\times$256$\times$100, shown in~\figurename~\ref{fig:testimages3d}~(a), to compare the fastest algorithms from Section~\ref{sec:2dtests}. This was done by computing grey-weighted distance transforms from manually placed seeds in a gradient magnitude image. The gradient magnitude image can be considered a low complexity image with large uniform areas and has high spatial correlation between voxels, i.e., properties similar to the BALL image in Section~\ref{sec:2dtests}. Two transforms were calculated, one from seeds placed outside the liver and one from seeds placed inside the liver, and the average runtime for each setup is shown in~\tablename~\ref{tab:runtimes_3dtransform}.

\begin{table*}
\begin{center}
\caption{Runtime (in seconds) for grey-weighted distance transforms on the CT and CE-MRA images. The best times are marked with black colour, times within 10\% of the best with dark grey, and times within 40\% of best with light grey.}
\label{tab:runtimes_3dtransform}
\medskip
\footnotesize{
\begin{tabular}{l*{7}{r}}
\hline
\noalign{\smallskip}
 & \multicolumn{3}{c}{CT}  & & \multicolumn{3}{c}{CE-MRA} \\
\cline{2-4}
\cline{6-8}
\noalign{\smallskip}
 & GRAYMAT & DOCS & WDOCS  & & GRAYMAT & DOCS & WDOCS \\
\hline
\noalign{\smallskip}
H        &    50.86 &    \forty{54.13} &    86.28 & &   257.97 &   \forty{183.30} &   385.54 \\ 
H\sub{A} &    \forty{37.49} &    \forty{49.23} &    \forty{54.77} & &   \forty{152.28} &   \best{163.12} &   \ten{232.13} \\ 
F &    55.62 &    62.25 &   104.59 & &   306.65 &   \forty{218.62} &   459.10 \\ 
F\sub{A} &    \forty{39.36} &    \forty{54.31} &    \forty{65.54} & &   \forty{169.56} &   188.18 &   \forty{263.24} \\ 
D\sub{L} &        - &    \ten{44.70} &        - & &        - &   \forty{180.81} &        - \\ 
D\sub{F} &        - &    \ten{44.48} &        - & &        - &   \forty{189.98} &        - \\ 
D\sub{LA} &        - &   563.30 &        - & &        - &  6067.59 &        - \\ 
U\sub{L} &    \forty{37.69} &    \ten{43.81} &    \forty{61.84} & &   191.38 &   \forty{183.69} &   \forty{278.04} \\ 
U\sub{F} &    \forty{37.56} &    \ten{42.52} &    \forty{60.84} & &   193.33 &   \forty{190.76} &   \forty{281.31} \\ 
U\sub{LA} &  2111.36 &   541.08 &   287.86 & &   270.88 &  6603.72 &  3012.72 \\ 
HH        &    \forty{38.45} &    \ten{44.22} &    69.22 & &   195.96 &   \forty{184.31} &   315.37 \\ 
HH\sub{A} &    \best{31.84} &    \best{42.19} &    \best{49.03} & &   \best{127.88} &   \ten{167.81} &   \best{212.52} \\ 
\hline
\end{tabular}
}
\end{center}
\end{table*}

Here it becomes apparent that the Dial queue and Untidy queue do not work well with the pointer array since updating a value is associated with a search through a list, which can be quite large due to the limited spread discussed in Section~\ref{sec:costspread}. For any of the grey-weighted distance transforms on the gradient magnitude of a CT image, the fastest algorithm is the hierarchical heap with the pointer array, but only marginally so for the DOCS definition. It is also noteworthy that the binary heap with pointer array performs better than the Untidy queue for both the GRAYMAT and WDOCS definitions.

\subsubsection{CE-MRA}

Here we used the 'lower thorax and abdomen' subvolume with size 512$\times$512$\times$154, shown in~\figurename~\ref{fig:testimages3d}~(b), to compare the algorithms used in Section~\ref{sec:3dtests_ct}. This was done by computing one distance transform tracing a route through the iliac running down the left leg and terminating at the bottom of the image. The runtime for each setup is shown in~\tablename~\ref{tab:runtimes_3dtransform}.

The runtimes for the CE-MRA dataset show results similar to the CT dataset. The fastest algorithm is the hierarchical heap with pointer array, and the runner up is the binary heap with pointer array.

\subsubsection{Dynamic vs static priority queues}\label{sec:static}

In addition to the dynamic versions of the Dial and Untidy queue that use STL List containers, both priority queues were also implemented using static arrays to give an indication of the tradeoff between static and dynamic priority queues. As mentioned in Section~\ref{sec:technicaldetails}, a static array implementation has a high programming complexity compared to using the STL List. It also requires at least twice as much memory as the distance map while a dynamic priority queue only requires enough space to accommodate the front wave (sometimes referred to as the 'narrow band'). \tablename~\ref{tab:runtimes_3dtransform_static} lists the runtimes for the binary heap with pointer array (H\sub{A}), Dial queue (D\sub{L}), Dial queue using static array (D\sub{SL}), Untidy queue (U\sub{L}), Untidy queue using static array (U\sub{SL}), and hierarchical heap with pointer array (HH\sub{A}). The static array implementations of the Dial and Untidy queue show an increase in performance by \mbox{$\sim$ 5-33\%} compared to their dynamic implementations but only \mbox{$\sim$ 0-11\%} compared to the fastest algorithms with dynamic priority queues. Note that when calculating the DOCS transform on the CE-MRA, the static array implementations ran slower than both the binary heap with pointer array and the hierarchical heap with pointer array. Considering the drawbacks with implementing static array priority queues, these runtimes indicate that dynamic priority queues can provide a fair tradeoff between performance and programming complexity/memory usage.

\begin{table*}
\begin{center}
\caption{Runtime (in seconds) for priority queues using static arrays (grey rows) compared to using dynamically allocated queues.}
\label{tab:runtimes_3dtransform_static}
\medskip
\footnotesize{
\begin{tabular}{l*{7}{r}}
\hline
\noalign{\smallskip}
 & \multicolumn{3}{c}{CT}  & & \multicolumn{3}{c}{CE-MRA} \\
\cline{2-4}
\cline{6-8}
\noalign{\smallskip}
 & GRAYMAT & DOCS & WDOCS  & & GRAYMAT & DOCS & WDOCS \\
\hline
\noalign{\smallskip}
H\sub{A} & 37.49 & 49.23 & 54.77 & & 152.28 & 163.12 & 232.13 \\ 
D\sub{L} & - & 44.70 & - & & - & 180.81 & - \\ 
D\sub{SL} \cfill & \cfill - & \cfill 41.62 & \cfill - & \cfill & \cfill - & \cfill 169.10 & \cfill - \\ 
U\sub{L} & 37.69 & 43.81 & 61.84 & & 191.38 & 183.69 & 278.04 \\ 
U\sub{SL} \cfill & \cfill 30.34 & \cfill 41.67 & \cfill 44.36 & \cfill & \cfill 127.38 & \cfill 171.11 & \cfill 191.77 \\ 
HH\sub{A} & 31.84 & 42.19 & 49.03 & & 127.88 & 167.81 & 212.52 \\ 
\hline
\end{tabular}
}
\end{center}
\end{table*}

\section{Conclusion}
\label{sec:conclusion}

The performance variations do not motivate a different choice of algorithm for different grey-weighted distance definitions, i.e., if an algorithm performs well for one definition, it is likely to perform well for the other two. The same holds for the difference in image properties. For 2D images, the hierarchical heap using pointer array shows to be the best choice. The same choice also proves to be the best for 3D images ($\sim$ 0-16\% faster than the second best). However, if memory is a critical issue, i.e., if the pointer array helper structure is not an option, then the Dial queue is the best option if we work with integer costs, and the Untidy queue if we work with real valued costs. It is noteworthy that the popular binary heap with pointer array performs quite well compared to the more sophisticated priority queues ($\sim$ 0-19\% slower than the fastest algorithm).


\section{Acknowledgements}

The Department of Radiology, Uppsala University Hospital, is acknowledged for providing the CT and CE-MRA datasets. Stina Svensson, Robin Strand, Cris Luengo, and Filip Malmberg, Centre for Image Analysis, Uppsala, Sweden, are acknowledged for scientific support. Magnus Gedda is financially supported by the Swedish Research Council (project 621-2005-5540).










\end{document}